\documentclass{elsarticle}

\usepackage[margin=1.35in]{geometry}
\usepackage{amsmath, amsfonts, amssymb}
\usepackage[hidelinks]{hyperref}
\usepackage[capitalize]{cleveref}
\usepackage{natbib}
\usepackage{booktabs}
\usepackage{caption, subcaption}
\usepackage{xcolor}
\usepackage{algorithm}
\usepackage{algpseudocode}
\usepackage{multirow}
\usepackage{csquotes}
\usepackage{graphicx}
\usepackage{soul}

\usepackage{amsthm}
\newtheorem{theorem}{Theorem}
\newtheorem{lemma}{Lemma}
\newtheorem{remark}{Remark}

\newtheorem{assumption}{Assumption}

\newtheorem*{definition*}{Definition}

\newcommand{\dd}{\mathrm{d}}
\newcommand{\response}{\color{black}}

\begin{document}

\title{Neural Operator-Based Nonlinear Nudging for Chaotic Dynamical Systems}

\author[1]{Jaemin Oh}
\ead{jaeminoh.math@gmail.com}

\author[2]{Jinsil Lee}
\ead{jl74942@snu.ac.kr}

\author[2,3]{Youngjoon Hong\corref{cor2}}
\ead{hongyj@snu.ac.kr}

\cortext[cor2]{Corresponding author}

\affiliation[1]{organization={Division of Applied Mathematics, Brown University}, addressline={170 Hope Street}, city={Providence}, postcode={RI 02906}, country={USA}}
\affiliation[2]{organization={Research Institute of Mathematics, Seoul National University}, addressline={1 Gwanak-ro, Gwanak-gu}, city={Seoul}, postcode={08826}, country={Republic of Korea}}
\affiliation[3]{organization={Department of Mathematical Sciences, Seoul National University}, addressline={1 Gwanak-ro, Gwanak-gu}, city={Seoul}, postcode={08826}, country={Republic of Korea}}

\begin{abstract}
Nudging is an empirical data assimilation technique that incorporates an observation-driven control term into the model dynamics. The trajectory of the nudged system approaches the true system trajectory over time, even when the initial conditions differ. For linear state space models, such control terms can be derived under mild assumptions. However, designing effective nudging terms becomes significantly more challenging in the nonlinear setting.
In this work, we propose neural network nudging, a data-driven method for learning nudging terms in nonlinear state space models. We establish a theoretical existence result based on the Kazantzis--Kravaris--Luenberger observer theory. The proposed approach is evaluated on three benchmark problems that exhibit chaotic behavior: the Lorenz 96 model, the Kuramoto--Sivashinsky equation, and the Kolmogorov flow.
\end{abstract}


\maketitle

\section{Introduction}
Data assimilation combines an imperfect predictive model with sparse and noisy observations in order to estimate the true state of a physical system.  The procedure is indispensable in numerical weather prediction, where sensitive dependence on initial conditions amplifies small errors and destroys forecast skill~\citep{lorenz1963deterministic}.  Modern atmospheric and oceanic models resolve phenomena across a wide range of scales and may involve billions of degrees of freedom, so any feasible assimilation strategy must balance statistical rigor with computational efficiency.  Outside the geosciences, similar needs arise in robotics~\citep{abidi1992data}, target tracking~\citep{gordon1993novel}, and biomedical monitoring~\citep{kim2024wearable}, which demonstrates the broad practical relevance of the discipline.

Two principal frameworks dominate current practice in data assimilation. Variational methods, exemplified by four-dimensional variational assimilation (4D-Var), pose the estimation task as a constrained optimization problem, where the cost function quantifies the discrepancy between the model trajectory and the observations. Sequential methods, such as the Kalman filter and its ensemble-based extensions, instead propagate a collection of states forward in time and update them whenever new observations become available.  While both approaches are grounded in established statistical theory, their optimality faces challenges when applied to realistic settings.  Implementing 4D-Var necessitates the construction of a tangent–linear and adjoint model, which is labor-intensive, prone to errors, and incurs a memory cost that scales with the assimilation window length. Furthermore, the resulting optimization problem is non-convex, meaning that solvers may converge to suboptimal local minima without a high-quality initial guess or proper regularization~\citep{nocedal1999numerical,oh2026effects}.
Ensemble-based filters do not require tangent-linear or adjoint models, as they directly approximate error covariances from an ensemble of forecasts.
However, the ensemble size required to accurately represent covariances and to mitigate sampling errors grows rapidly with system dimension and increasing nonlinearity~\citep{bellman1957dynamic}.  Although inflation and localization heuristics can reduce this burden, they introduce additional hyperparameters that must be carefully tuned for each application.  Particle filters, while theoretically capable of approximating the full Bayesian posterior~\citep{law2015data}, are prone to weight degeneracy in high-dimensional settings, rendering them impractical for most operational scenarios.

Nudging, also referred to as Newtonian relaxation, offers a practical alternative that is simple to implement and empirically robust even in the presence of significant model error~\citep{lei2015nudging,clark2020synchronization,carlson2024super}. The core idea is to incorporate a feedback control term into the prognostic equation, allowing the model state to gradually align with the available observations. For fully linear and observable systems, a fixed gain matrix can be constructed to ensure exponential convergence of the nudged trajectory to the true state. However, real-world systems are rarely both linear and fully observable, which often necessitates heuristic tuning of gain parameters. Gains derived from linearizations may perform poorly and can even introduce instability when the underlying dynamics are strongly nonlinear. Nonlinear observer theory, particularly the Kazantzis--Kravaris--Luenberger (KKL) framework, provides a theoretical basis for handling nonlinearity by constructing a global coordinate transformation that renders the system observable in the transformed space~\citep{kazantzis1998nonlinear}.  Nevertheless, computing such a transformation remains computationally challenging in practice. These developments underscore the need for nudging methods that can automatically adapt to nonlinear behavior without manual gain design.

We address this need with \emph{neural network nudging} (NNN), a data-driven methodology that replaces manually tuned gain terms with a learnable operator parameterized by a neural network. Instead of modifying the underlying model or constructing a surrogate, our approach directly learns the feedback control term. To enable the application to high-dimensional spatial fields, we employ a modified deep neural operator (DNO)~\citep{lu2019deeponet}. The network is trained offline using pairs of model states and synthetic observations generated from the known physical system. Once trained, the method incurs negligible computational overhead during inference, as it requires only a single forward pass at each time step.

The proposed framework, NNN, contributes to the literature in three main aspects. First, it reformulates the design of nudging gains as a supervised learning problem in operator form, removing the need for manual tuning and allowing data to determine a nonlinear feedback law. Second, it establishes a theoretical foundation by proving an existence result that links NNN to the KKL observer. Specifically, the result shows that if a suitable invertible KKL transformation exists, then there also exists a neural operator that guarantees state synchronization. Third, the framework is validated empirically on a sequence of benchmark problems of increasing dynamical complexity. The Lorenz 96 system~\citep{lorenz1998optimal} serves as a canonical low-dimensional chaotic model. The Kuramoto--Sivashinsky equation captures instability and dissipation in one spatial dimension. The two-dimensional incompressible Navier--Stokes equations with Kolmogorov forcing provide a stringent test. In most cases, NNN achieves lower analysis error than linear nudging.

Previous studies that combine machine learning with data assimilation have mostly focused on augmenting existing variational or ensemble pipelines. Examples include learned observation operators for 4D-Var~\citep{frerix2021variational}, neural reparametrization for 4D-Var~\citep{oh2026effects}, neural pseudo-inverses for constraint satisfaction~\citep{filoche2023learning}, and score-based approaches for trajectory estimation~\citep{rozet2023score,bao2024score}.  Ensemble-free neural filters have also been proposed~\citep{bocquet2024accurate}.  To the best of our knowledge, only a few studies have explored the connection between nudging and neural networks, despite their conceptual alignment with neural ordinary differential equations~\citep{chen2018neural}. \citet{antil2024data} proposed fitting a residual neural network (ResNet) to a nudged model, while \citet{antil2024ninns} introduced a method for nudging a ResNet trained on time-series data. \citet{barthel2023machine} proposed to use a nudged dataset for training downscaling models. Our approach differs fundamentally from these approaches. Rather than applying nudging to models or training a network for a nudged model, we represent the nudging term itself as a neural network. This formulation aims to overcome the limitations of traditional linear nudging by enabling flexible and data-adaptive correction.

The remainder of the paper is organized as follows. \Cref{sec-da} reviews classical assimilation techniques and revisits nudging theory. \Cref{sec-nnn} details the NNN formulation and presents the existence theorem. \Cref{sec-results} reports numerical results, and \Cref{sec-conclusion} summarizes the main findings and outlines future directions.

\section{Data assimilation}\label{sec-da}
\begin{figure}[!t]
    \centering
    \includegraphics[width=1\linewidth]{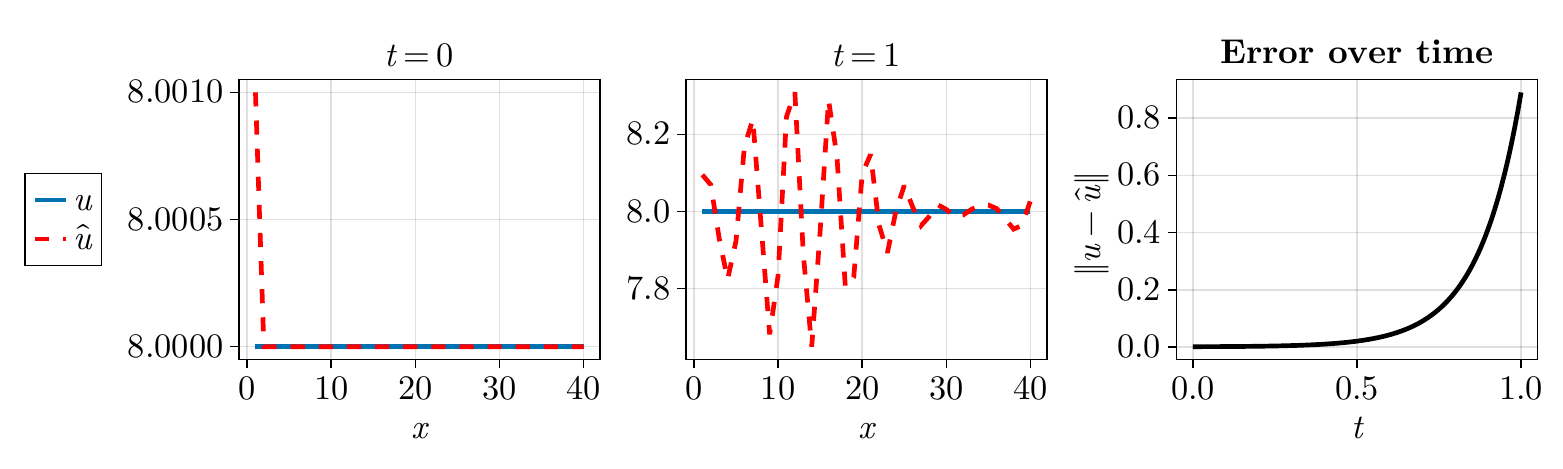}
    \caption{Chaotic system's sensitivity to initial conditions. The left two panels present solution snapshots at \(t=0\) and \(t=1\), respectively. The rightmost panel shows the rapid growth of \(\ell_2\) distance between two solutions in time. The underlying model is the Lorenz96 system (see \cref{eq-L96}).}
    \label{fig-chaos}
\end{figure}

In 1961, Edward Lorenz discovered that truncating the last few decimal points of the initial condition produced a totally different forecast in his numerical weather model.\footnote{https://www.aps.org/archives/publications/apsnews/200301/history.cfm} In a later account~\citep{lorenz1993essence} he wrote:
\begin{displayquote}
    \emph{The initial round-off errors were the culprits; they were steadily amplifying until they dominated the solution. In today’s terminology, there was chaos.}
\end{displayquote}
This indicates the extreme sensitivity of the model to initial conditions. \Cref{fig-chaos} illustrates an example of the sensitivity; a discrepancy of \(10^{-3}\) in the first component of the initial condition contaminates the solution at \(t = 1\). Such sensitivity renders exact numerical prediction extremely challenging~\citep{lorenz1963deterministic,trefethen1993hydrodynamic}, a characteristic known as deterministic chaos.

\begin{figure}[!ht]
    \centering
    \includegraphics[width=0.5\linewidth]{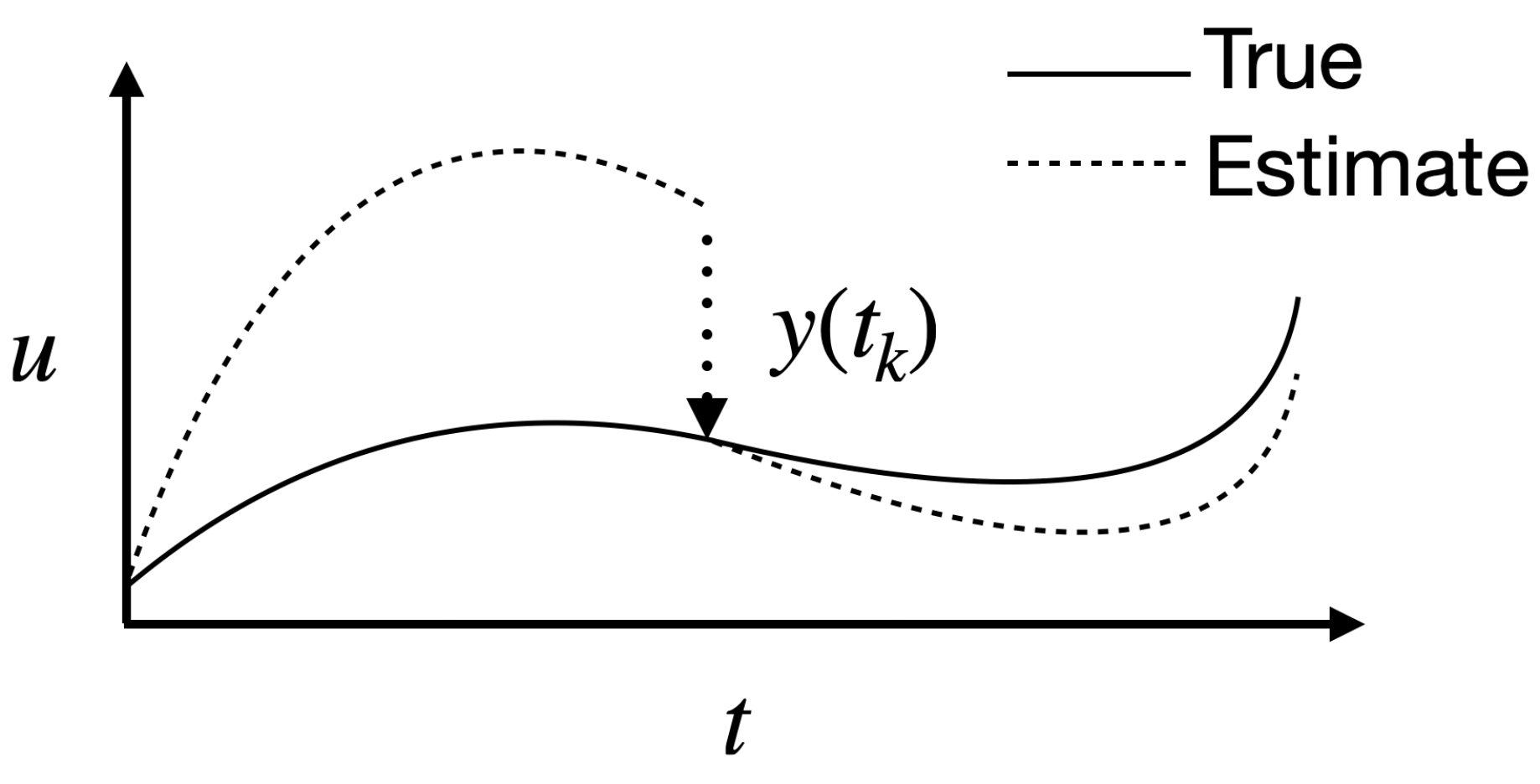}
    \caption{Graphical illustration of data assimilation. \(y(t_k)\) denotes an observation at a time point \(t_k\). Data assimilation utilizes \(y(t_k)\) to correct the deviation between the true state and the estimated state caused by the initial discrepancy.}
    \label{fig-assimilation}
\end{figure}

For chaotic systems, one practical approach to improve prediction accuracy is to restart the simulation from an initial state that is closer to reality.  Observations, although incomplete and noisy, provide partial information about the true state and can be used for this purpose.  Data assimilation combines these observations with model predictions to estimate the underlying state more accurately.  As shown in \Cref{fig-assimilation}, the discrepancy between the model trajectory and the actual system can be reduced by adjusting the state toward the available observations, after which subsequent predictions inherit the improved accuracy.  Assimilation algorithms differ in the metrics they minimize and in the optimization strategies they employ.  Kalman filters and their smoother variants seek to minimize the posterior error covariance, whereas variational methods minimize a cost functional that measures the joint mismatch between model and data over a time window.  Mathematical treatments of these approaches can be found in~\citet{law2015data}.

Many time-dependent phenomena in science and engineering are governed by partial differential equations (PDEs). When discretized in space -- using finite difference, finite volume, or spectral methods -- these PDEs are reduced to systems of ordinary differential equations that describe the temporal evolution of the discretized state~\citep{trefethen1996finite,leveque2002finite,hesthaven2007spectral}. This process yields high-dimensional dynamical systems that serve as the basis for numerical simulation and data assimilation. Accordingly, we restrict our attention to initial value problems of the form
\begin{equation}\label{eq-model}
    \frac{\dd u(t)}{\dd t} = \mathcal{F}\bigl(u(t)\bigr),
\end{equation}
where \(u(t)\) represents the discretized system state at time \(t\), and \(\mathcal{F} \colon \mathbb{R}^d \to \mathbb{R}^d\) encodes the resolved dynamics.  Observations are modeled as
\begin{equation}\label{eq-observation}
    y(t) = \mathcal{H}\bigl(u(t)\bigr) + \varepsilon(t),
\end{equation}
where \(\mathcal{H} \colon \mathbb{R}^d \to \mathbb{R}^p\) is a continuous observation operator and \(\varepsilon(t)\) denotes measurement noise, typically assumed to be white. The pair \cref{eq-model,eq-observation} defines a continuous-time state-space model, which forms the basis for the analysis throughout this work.

\subsection{Nudging}
Nudging adds an observation-driven relaxation term to \cref{eq-model} to steer the model state toward the observed data. This approach is a type of state observer that augments the model dynamics with a control term based on observational feedback. A classical formulation is provided by the Luenberger observer~\citep{luenberger1964observing,luenberger1966observers,luenberger1971introduction}. Assuming both \(\mathcal{F}\) and \(\mathcal{H}\) are linear operators (denoted by \(\mathcal F_L\) and \(\mathcal H_L\), respectively) and that the measurement noise \(\varepsilon(t)\) is absent, the observer equation takes the form
\begin{equation}\label{eq-linear-observer}
    \frac{\dd \hat{u}(t)}{\dd t} = \mathcal{F}_L \hat{u}(t) + \mathcal{G}_L(y(t) - \mathcal{H}_L \hat{u}(t)),
\end{equation}
where \(\hat{u}(t)\) is the estimate of the system state, evolved from an initial guess \(\hat{u}(0)\). The term \(\mathcal{G}_L(y - \mathcal{H}_L \hat{u})\) acts as a feedback control, driving the estimated trajectory toward the true state \(u(t)\).  To analyze the convergence, let \(E(t) = \hat{u}(t) - u(t)\).  Subtracting the true dynamics in \cref{eq-model} from \cref{eq-linear-observer} and using \(y(t) = \mathcal{H}_L u(t)\) yields the error equation
\[
    \frac{\dd E(t)}{\dd t} = (\mathcal{F}_L - \mathcal G_L \mathcal H_L)E(t).
\]
If all eigenvalues of \(\mathcal{F}_L - \mathcal G_L \mathcal H_L\) are negative, then the error decays exponentially:
\begin{equation}\label{eq-luenberger-error}
    \|E(t)\| = O\bigl( e^{\lambda_1 t} \bigr), \qquad \lambda_1 < 0,
\end{equation}
where \(\lambda_1\) is the maximum eigenvalue of \(\mathcal{F}_L - \mathcal G_L \mathcal H_L\). It is known that if the pair \((\mathcal F_L, \mathcal H_L)\) satisfies the observability condition, then there exists a gain matrix \(\mathcal G_L\) such that all eigenvalues of \(\mathcal F_L - \mathcal{G}_L \mathcal H_L\) are negative real numbers~\citep[Lemma~1]{luenberger1971introduction}.

In contrast to Kalman filters and variational methods, which are derived from probabilistic principles and typically involve the minimization of posterior variance or a likelihood-based cost functional, nudging is an empirical approach.  Its main advantage lies in its simplicity and low computational cost, as it avoids matrix inversion, ensemble propagation, or adjoint computations.  However, the classical theory is limited to the linear and noise-free setting. When either \(\mathcal{F}\) or \(\mathcal{H}\) is nonlinear, or when the observations include noise, the linear analysis (\cref{eq-luenberger-error}) no longer applies directly. In such cases, designing effective nudging strategies becomes problem-dependent and more challenging. To address this, we consider a generalization of the linear observer (\cref{eq-linear-observer}) to the nonlinear case:
\begin{equation}\label{eq-observer}
    \frac{\dd \hat{u}(t)}{\dd t} = \mathcal{F}\bigl(\hat{u}(t)\bigr) + \mathcal{G}\bigl(\hat{u}(t), y(t)\bigr)
\end{equation}
where \(\mathcal{G}\) is a nonlinear feedback term. In the following section, we introduce a data-driven framework for learning \(\mathcal{G}\).

\section{Neural network nudging}\label{sec-nnn}
In this section, we introduce \emph{neural network nudging} as a method for nudging general nonlinear state–space models. The key idea is to replace the unknown feedback term \(\mathcal{G}\) in \cref{eq-observer} with a neural network \(\mathcal{G}_\theta\) parameterized by \(\theta\),
\begin{equation}\label{eq-nnn}
    \frac{\dd \tilde u(t)}{\dd t} = \mathcal{F}\bigl(\tilde u(t)\bigr) + \mathcal{G}_\theta\bigl(\tilde u(t), y(t)\bigr),
\end{equation}
where \(\tilde u(t)\) denotes the nudged state. The parameters \(\theta\) are learned by minimizing a discrepancy loss between nudged trajectories and ground truth data, using standard backpropagation. This formulation leverages the expressiveness of neural networks to represent complex feedback laws and is expected to improve upon linear nudging, particularly in nonlinear regimes. The theoretical motivation stems from the universal approximation property of neural operators, which suggests that suitably parameterized networks can emulate the convergence behavior observed in linear observers such as \cref{eq-luenberger-error}. We explore this connection in detail below, where we establish an existence result (\Cref{thm-uat-kkl}) by combining the KKL observer framework (\Cref{thm-kkl}) with the universal approximation theorem for deep neural operators (\Cref{thm-uat-operator}). The empirical benefits of NNN over classical linear nudging will be validated through numerical experiments in \Cref{sec-results}.

\subsection{Existence of neural network nudging terms}
\paragraph{KKL observer theory}
\citet{kazantzis1998nonlinear} extended the Luenberger observer framework to nonlinear state-space models, now known as the KKL observer. The key idea is to construct a nonlinear coordinate transformation \(T\colon \mathbb{R}^d \to \mathbb{R}^m\) such that the transformed state \(z(t) = T(u(t))\) evolves under a linear system up to the observation term:
\begin{equation}\label{eq-kkl}
    z(t) = T(u(t)), \qquad \frac{\dd z(t)}{\dd t} = Az(t) + By(t), \qquad z(0) = T(u(0)),
\end{equation}
where \(A \in \mathbb{R}^{m \times m}\) and \(B \in \mathbb{R}^{m \times p}\) are matrices chosen such that the pair \((A, B)\) is controllable.\footnote{That is, the controllability matrix \([B \ AB \ \dots\ A^{m-1}B]\) has full rank \(m\).} If \(\hat{z}(0) = z(0) + \varepsilon\), then the corresponding trajectory \(\hat{z}(t)\) converges to \(z(t)\) as \(t \to \infty\). Here, we use the notation \(\hat{\cdot}\) to denote trajectories starting from perturbed initial conditions. The map \(T\) linearizes the nonlinear dynamics (up to the observation term) in the transformed coordinates. This global transformation approach was first proposed by \citet{gilbert1984approach} for observer design.

To determine \(T\), we consider a necessary condition for its existence. We shall drop \(t\) if there is no confusion. By substituting \(z = T(u)\)  into~\cref{eq-kkl} and applying the chain rule, one obtains the PDE
\begin{equation}\label{eq-kkl-pde}
    \partial T(u) \mathcal F(u) = AT(u) + B\mathcal{H}(u), \qquad T(0) = 0,
\end{equation}
where \(\partial T(u)\) denotes the Jacobian of \(T\) at \(u\). The condition \(T(0) = 0\) fixes a unique solution up to translation.
{\response In \citet[Theorem 2]{kazantzis1998nonlinear}, the authors assumed the existence of a \emph{global} solution to~\cref{eq-kkl-pde}. Following their framework, we formalize this requirement in the following assumption.
\begin{assumption}\label{assumption-T}
There exists a controllable pair \((A,B)\), where \(A\) is diagonalizable and its eigenvalues are negative, such that~\cref{eq-kkl-pde} admits an invertible and continuously differentiable solution \(T: \mathbb{R}^d \rightarrow \mathbb{R}^d\).
\end{assumption}
\begin{remark}
The condition that \(A\) is diagonalizable can be relaxed by considering its Jordan canonical form. For simplicity, however, we restrict our discussion to the case where \(A\) is diagonalizable. The existence and regularity of solutions to~\cref{eq-kkl-pde} under weaker assumptions are studied in detail by \citet{brivadis2023further}.
\end{remark}
}
Under this assumption, we establish the asymptotic convergence of the observer. Unless explicitly stated otherwise, \(\|\cdot\|\) denotes the standard Euclidean distance.
\begin{theorem}\label{thm-kkl}
Let Assumption~\ref{assumption-T} hold and consider \(\varepsilon (t)=0\). Then, the dynamical system
\begin{equation}\label{eq-KKL}
\frac{\mathrm{d} \hat{u}}{\mathrm{d} t} = \mathcal{F}(\hat{u}) + \left[\partial T(\hat{u})\right]^{-1} B\big(y - \mathcal{H}(\hat{u})\big)
\end{equation}
is an identity observer for the original system~\eqref{eq-model}, satisfying
\[
    \lim_{t \to \infty} \|T(\hat{u}(t)) - T(u(t))\| = 0.
\]
\end{theorem}
\begin{proof}
    See \cite[Theorem~2]{kazantzis1998nonlinear}.
\end{proof}

Motivated by the formulation in \cref{eq-KKL}, we expect the feedback term \(\mathcal{G}_\theta\) to depend explicitly on the current estimate \(\hat{u}\).  In particular, note that the mapping
\[
    \mathcal{G}(\hat{u})(y) := \partial T(\hat{u})^{-1} B\bigl( y - \mathcal{H}(\hat{u}) \bigr)
\]
defines an operator. In what follows, we approximate this operator using a deep neural operator:
\[
    \mathcal{G}_\theta(\hat{u})(y ) \approx \partial T(\hat{u})^{-1} B\bigl( y - \mathcal{H}(\hat{u}) \bigr).
\]

\paragraph{Deep neural operators}
Deep neural operators, originally introduced as deep operator networks~\citep{lu2019deeponet}, are neural architectures designed to approximate continuous nonlinear operators defined on compact subsets of Banach spaces. The universal approximation theorem for operators~\citep{chen1995universal,lu2019deeponet} provides their theoretical justification.

A deep neural operator takes the form
\[
    \mathcal G_\theta(u)(y) = \sum_{c=1}^C \mathcal{B}_c(u) \, \mathcal{T}_c(y),
\]
where \(\mathcal{B}_c\) and \(\mathcal{T}_c\) denote the outputs of the \emph{branch} and \emph{trunk} networks, respectively. The branch network \(\mathcal{B}\) processes the input function \(u\) and outputs coefficients, while the trunk network \(\mathcal{T}\) provides basis functions evaluated at the target location \(y\). This separation allows the network to learn mappings between infinite-dimensional function spaces in a flexible and computationally efficient manner.

The following theorem, established by \citet{chen1995universal} and later revitalized in the context of deep learning by \citet{lu2019deeponet}, shows that deep neural operators can approximate (nonlinear) continuous operators:
\begin{theorem}[Universal approximation theorem for operators]\label{thm-uat-operator}
    Let \(X\) be a Banach space, and let \(K_1 \subset X\), \(K_2 \subset \mathbb{R}^p\) be compact subsets. Let \(V \subset C(K_1)\) be a compact set, and let \(\mathcal{G} : V \rightarrow C(K_2)\) be a nonlinear continuous operator. Then, for each continuous non-polynomial activation function and for each \(\delta > 0\), there exists a deep operator network \(\mathcal{G}_\theta(u)(y)\) such that
    \[
        \sup_{u \in V,\, y \in K_2} \left| \mathcal{G}(u)(y) - \mathcal{G}_\theta(u)(y) \right| < \delta.
    \]
\end{theorem}
 It guarantees that DNOs can uniformly approximate any continuous operator defined on compact subsets of Banach spaces. As such, they offer a powerful framework for learning mappings between function spaces, particularly those arising in the context of PDEs and dynamical systems.

We now verify that the operator \(\mathcal{G}(\hat{u})(y) = \partial T(\hat{u})^{-1} B \bigl( y - \mathcal{H}(\hat{u}) \bigr)\) satisfies the conditions required by \Cref{thm-uat-operator}. For this purpose, we impose the following assumptions:
\begin{assumption}\label{assumption-2}
    There exists a compact set \(V\) such that the trajectory \(u(t; u_0)\), corresponding to the solution of \cref{eq-model} with initial condition \(u_0 \in \mathbb{R}^d\), satisfies \(u(t; u_0) \in V\) for all \(t > 0\) and all \(u_0\).
\end{assumption}

\begin{assumption}\label{assumption-3}
    The Jacobian \(\partial T(u)\) is invertible for all \(u \in V\); that is, \(\det(\partial T(u)) \ne 0\) for all \(u \in V\).
\end{assumption}
Now, we have the following lemma:
\begin{lemma}\label{lemma-uat}
Under Assumptions \ref{assumption-T}, \ref{assumption-2}, \ref{assumption-3} and \(\varepsilon(t) = 0\), for any \(\delta >0\), there exists a DNO \(\mathcal G_\theta\) satisfying
\[
    \|\mathcal{G}_\theta(\hat{u})(y) - \partial T(\hat{u})^{-1}B\bigl(y - \mathcal H(\hat{u})\bigr)\| < \delta.
\]
\end{lemma}
\begin{proof}
The proof strategy involves verifying the conditions required by \Cref{thm-uat-operator}. First, by Assumption~\ref{assumption-2}, the set of states $u$ is compact. 
Next, since the observation operator $\mathcal{H}$ is continuous and the states $u$ belong to a compact set, the set of corresponding observations $y$ is also compact; explicitly, this observation set is given by $K_2 = \mathcal{H}(V)$, noting that $\varepsilon = 0$. Finally, we show that the operator $\mathcal{G}(u)(y) = \partial T(u)^{-1} B (y - \mathcal{H}(u))$ is continuous with respect to $u$. To this end, it suffices to show the continuity of the mapping $u \mapsto \partial T(u)^{-1}$. 
This follows directly from Assumption~\ref{assumption-3} and the inverse function theorem, as $T^{-1}$ is $C^1$, and therefore $\partial T(u)^{-1} = (\partial T^{-1})(T(u))$ is a composition of continuous functions. Since all the conditions of \Cref{thm-uat-operator} are satisfied, the proof is complete.
\end{proof}

\begin{theorem}\label{thm-uat-kkl}
    {\response
    Let us assume Assumptions \ref{assumption-T}, \ref{assumption-2}, \ref{assumption-3}.
    Let \(u(t) \in V\) be the solution to \cref{eq-model} from an initial condition \(u(0) = u_0\). Let \(\tilde{u}(0) = \tilde{u}_0 \in \mathbb{R}^d\) be another initial condition.} Under \(\varepsilon(t) = 0\), for any \(\delta > 0\), there exists a neural network parameter \(\theta\) and time \(\tau \geq 0\) such that the solution \(\tilde{u}(t)\) to \cref{eq-nnn} satisfies
    \[
        \| T(\tilde{u}(t)) - T(u(t)) \| \le \delta \qquad \text{for all } t \ge \tau.
    \]
\end{theorem}
Before proceeding to the proof, we first state the following lemma, which will be used in the argument.
\begin{lemma} \label{lemma-asymptotic-bound}
    {\response Let \(E: (0, \infty) \rightarrow \mathbb{R}^d\) be differentiable. Let \(A \in \mathbb{R}^{d\times d}\) be a diagonalizable matrix whose eigenvalues are negative.} Suppose \(\left\|\dfrac{\dd E}{\dd t} - AE\right\| < \delta\). Then for sufficiently large \(t\), we have:
    \[
        \|E(t)\| \le \frac{C\delta}{-\lambda_1},
    \]
    where \(\lambda_1\) is the largest eigenvalue of \(A\), and \(C = \|Q\|\|Q^{-1}\|\) for the diagonalization \(A = Q \Lambda Q^{-1}\).
\end{lemma}
\begin{proof}
Let \(R = \dfrac{\dd E}{\dd t} - AE\). Let \(V(t) = e^{-At}E(t)\). Then
\[
    \frac{\dd V(t)}{\dd t} = e^{-At} R(t).
\]
Integrating both sides from \(0\) to \(t\), we get:
\[
    V(t) = E(0) + \int_0^t e^{-As} R(s) \, \dd s,
\]
and multiplying both sides by \(e^{At}\) gives
\[
    E(t) = e^{At} E(0) + e^{At} \int_0^t e^{-As} R(s) \, \dd s.
\]
Taking norms and using the triangle inequality:
\[
    \|E(t)\| \le \|e^{At} E(0)\| + \left\|e^{At} \int_0^t e^{-As} R(s) \, \dd s \right\|.
\]
As \(t \to \infty\), the first term decays to zero. For the second term:
\begin{align*}
    \left\|e^{At} \int_0^t e^{-As} R(s) \, \dd s \right\|
    & {\response < \delta \int_0^t \|e^{A(t-s)}\| \, \dd s} \\
    &\le \delta C \int_0^t e^{\lambda_1 (t - s)} \, \dd s \\
    &= \frac{\delta C}{-\lambda_1} (1 - e^{\lambda_1 t}) \\
    &\le \frac{C \delta}{-\lambda_1}.
\end{align*}
\end{proof}

We are now ready to present the proof of \Cref{thm-uat-kkl}:
\begin{proof}
Let \(E = T(\tilde{u}) - T(\hat{u})\). We derive
\begin{align*}
    \frac{\dd E}{\dd t} 
    &= \partial T(\tilde{u}) \frac{\dd \tilde{u}}{\dd t} - \partial T(\hat{u}) \frac{\dd \hat{u}}{\dd t} \\
    &= \partial T(\tilde{u})\left[\mathcal{F}(\tilde{u}) + \mathcal{G}_\theta(\tilde{u})(y)\right] \\
    &\quad - \partial T(\hat{u})\left[\mathcal{F}(\hat{u}) + \partial T(\hat{u})^{-1} B(y - \mathcal{H}(\hat{u}))\right] \\
    &= AT(\tilde{u}) + B\mathcal{H}(\tilde{u}) + \partial T(\tilde{u}) \mathcal{G}_\theta(\tilde{u})(y) \\
    &\quad - AT(\hat{u}) - B\mathcal{H}(\hat{u}) - By + B\mathcal{H}(\hat{u}) \\
    &= A E + \underbrace{\partial T(\tilde{u}) \left[\mathcal{G}_\theta(\tilde{u})(y) - \partial T(\tilde{u})^{-1} B(y - \mathcal{H}(\tilde{u}))\right]}_{=: R}.
\end{align*}
The first equality comes from the chain rule, the second equality comes from \cref{eq-nnn} and \cref{eq-KKL}, the third equality comes from \cref{eq-kkl-pde}, and the last equality comes from reordering and factoring out \(\partial T(\tilde u)\). {\response Now the error satisfies
\[
    \frac{\dd E}{\dd t} = AE + R.
\]
Since \(\partial T(\tilde{u})\) is bounded (the range of \(u \mapsto \partial T(u)\) is a compact subset of \(\mathbb R^d\), hence bounded), and by \Cref{lemma-uat}, we can choose \(\theta\) such that \(\|R\| < -\lambda_1 \delta / C\), where \(\lambda_1\) is the largest eigenvalue of \(A\), and \(C = \|Q\| \|Q^{-1}\|\) with \(A = Q \Lambda Q^{-1}\).} Applying \Cref{lemma-asymptotic-bound}, we conclude that \(\|E(t)\| < \delta\) for all sufficiently large \(t\).

Finally, since \(\|T(\hat{u}(t)) - T(u(t))\| \to 0\) as \(t \to \infty\), we obtain the desired result.
\end{proof}
\begin{remark}
    Since \(T^{-1}\) is continuous on a compact set, it is uniformly continuous. Therefore, \(\|\tilde u(t) - u(t)\|\) can be made arbitrarily small as well.
\end{remark}
\begin{remark}
We have used the KKL observer theory to show the theoretical existence of a NNN term \(\mathcal{G}_\theta\). Obtaining a numerical approximation of solutions to \cref{eq-kkl-pde} is another story and unknown in general.
\end{remark}
\begin{remark}
We acknowledge that Assumption~\ref{assumption-T} -- a globally invertible KKL transformation \(T: \mathbb{R}^d \rightarrow \mathbb{R}^m\) with \(m = d\) -- is a restrictive idealization. For generic chaotic PDEs, ensuring global injectivity typically requires $m > d$, and the KKL transformation often renders the Jacobian $\partial T$ singular. Thus, the exact analytical cancellation in \Cref{thm-uat-kkl} serves primarily as an idealized mathematical scaffold to motivate the structural form of the nudging term. 
In practice, our data-driven neural network nudging framework circumvents the need to analytically compute this nonexistent inverse $\partial T^{-1}$. Because the system trajectories are confined to a compact global attractor $V$ (Assumption~\ref{assumption-2}), NNN only needs to learn a continuous, locally stabilizing feedback law restricted to this set. Supported by the Universal Approximation Theorem (\cref{thm-uat-operator}), NNN effectively bypasses the strict topological bottlenecks of classical observer theory by empirically learning this continuous operator directly from data.
\end{remark}

In applications, we have used a modified form of the deep neural operator:
\[
    \mathcal{G}_\theta(\tilde u)(y) = \sum_{c=1}^C \mathrm{MLP}_c(\tilde u) \mathrm{CNN}_c(y - \mathcal{H}(\tilde u)),
\]
where \(\mathrm{MLP}: \mathbb{R}^{d_u} \rightarrow \mathbb{R}^C\), and \(\mathrm{CNN}: \mathbb{R}^{d_y} \rightarrow \mathbb{R}^{d_u \times C}\), because we found that this form has performed well empirically. For the definition of \(\mathrm{MLP}\) and \(\mathrm{CNN}\), please see \ref{app-detail}.

\subsection{Training strategy}
In this section, we describe how to train the NNN term \(\mathcal{G}_\theta\). Following standard data-driven approaches, we first generate a dataset of true states by integrating the governing dynamics
\[
    \frac{\dd u}{\dd t} = \mathcal{F}(u)
\]
forward in time from an initial condition \(u(0)\).
We discretize the time interval $[0, T]$ with a uniform time step $\Delta t$, defining
\[
    t_k = k \Delta t, \quad k = 1, \dots, M,
\]
and collect \(M\) snapshots:
\[
    \{u(t_k) \mid k = 1, \dots, M\}.
\]
We use only the first \(M_\mathrm{tr}\) snapshots for training \(\mathcal{G}_\theta\) and reserve the remaining \(\{u(t_k) \mid k = M_\mathrm{tr} + 1, \dots, M\}\) for testing. Observations are generated as
\[
    y(t_k) = \mathcal{H}(u(t_k)) + \sigma \varepsilon(t_k), \qquad \varepsilon(t_k) \overset{\mathrm{iid}}{\sim} N(0, I),
\]
where \(\sigma\) is the noise scale.

We define the loss as the mean squared error between the predicted states \(\tilde{u}\) and the true states \(u\). Given the dataset, we consider the objective:
\begin{equation}\label{eq-loss}
L(\theta) = \frac{1}{IK} \sum_{i=0}^{I-1} \sum_{k=1}^{K} \left( \tilde{u}(t_{i,k}) - u(t_{i,k}) \right)^2,
\end{equation}
where we set \(M_\mathrm{tr} = I \cdot K\), use \(K\) as the unroll length, \(I\) is the quotient, and define \(t_{i,k} = (iK + k) \Delta t\). {\response This objective, evaluated on the dataset containing noise, enables \(\mathcal{G}_\theta\) to learn a nudging term that makes the trajectory close to the true dynamics.}

To predict the assimilated states efficiently, we use:
\begin{equation}\label{eq-correct}
    \tilde{u}(t_k) = \check u(t_k) + (t_k - t_{k-1}) \mathcal{G}_\theta\bigl(\check u(t_k)\bigr)\bigl(y(t_k)\bigr),
\end{equation} 
where
\begin{equation}\label{eq-predict}
    \check u(t_k) = \tilde{u}(t_{k-1}) + \int_{t_{k-1}}^{t_k} \mathcal{F}(\tilde{u}(s)) \, \dd s.
\end{equation}
In other words, we first evolve the state under \(\mathcal{F}\) until a new observation becomes available (\cref{eq-predict}), and then assimilate the observation via a single network evaluation (\cref{eq-correct}). This structure can be viewed as a first-order operator splitting method~\citep{pareschi2000implicit}.

For \cref{eq-loss}, both $K = 1$ and $K > 1$ are viable choices. The case $K = 1$ corresponds to a non-intrusive training setup that is computationally efficient, as it avoids backpropagation through the numerical solver. In contrast, using \(K>1\) requires backpropagation through numerical integrators, leading to higher memory demands.\footnote{Libraries such as \texttt{Diffrax}~\citep{kidger2021on} provide practical tools for adjoint-based backpropagation through ODE solvers.} Empirically, we observe that \(K>1\) improves accuracy despite the higher computational cost (see \Cref{tab-K-study}). 
We summarize the training procedure in \Cref{alg-training}.
\begin{algorithm}[!ht]
\caption{Training Procedure for \(\mathcal{G}_\theta\)}\label{alg-training}
\begin{algorithmic}
\Require
\begin{itemize}
    \item Observation interval \(\Delta t\)
    \item Training dataset \(\{u(t_k) \mid k = 1, \dots, M_\mathrm{tr}\}\)
    \item Unroll length \(K\)
    \item Observation noise variance \(\sigma^2\)
    \item Maximum optimization steps \(N_\mathrm{max}\)
    \item Initial network parameter \(\theta\)
\end{itemize}

\State Set \(\theta_0 = \theta\)
\State Generate noisy observations: \(y(t_k) = \mathcal{H}(u(t_k)) + \varepsilon(t_k)\), where \(\varepsilon(t_k) \sim \mathcal{N}(0, \sigma^2 I)\)

\For{\(n = 1\) to \(N_\mathrm{max}\)}
    \For{\(i = 0\) to \(I - 1\)}
        \State Initialize \(\tilde{u}(t_{i,0}) = u(t_{i,0}) + \varepsilon\), where \(\varepsilon \sim \mathcal{N}(0, \sigma^2 I)\)
        \For{\(k = 1\) to \(K\)}
            \State Compute \(\check u(t_{i,k}) = \tilde{u}(t_{i,k-1}) + \int_{t_{i,k-1}}^{t_{i,k}} \mathcal{F}(\tilde{u}(s)) \, \dd s\) \Comment{\cref{eq-predict}}
            \State Update \(\tilde{u}(t_{i,k}) = \check u (t_{i,k}) + \Delta t \, \mathcal{G}_{\theta_{n-1}}\bigl(\check u (t_{i,k})\bigr)\bigl(y(t_{i,k})\bigr)\) \Comment{\cref{eq-correct}}
        \EndFor
    \EndFor
    \State Update parameters: \(\theta_n = \theta_{n-1} - \eta \nabla_\theta L(\theta_{n-1})\) \Comment{Using \cref{eq-loss}}
\EndFor
\State \Return \(\theta_{N_\mathrm{max}}\)
\end{algorithmic}
\end{algorithm}

\section{Numerical results}\label{sec-results}
To assess the effectiveness of our algorithm, we perform numerical experiments on three dynamical systems: the Lorenz 96 model (\Cref{sec-L96}), the Kuramoto--Sivashinsky equation (\Cref{sec-KS}), and the Kolmogorov flow (\Cref{sec-KF}). These examples span a range of spatiotemporal complexity and are widely used benchmarks for data assimilation. In each set, we vary (i) the observation noise level and (ii) the proportion of observed variables to investigate how these factors affect the assimilation results. We compare our method with ensemble and ensemble transform Kalman filters (EnKF and ETKF, respectively). We additionally compare our method with a linear nudging baseline, implemented by specifying \(\mathcal{G}_\theta(u)(y) = \mathcal{G}_L(y - \mathcal{H}u)\) in \cref{eq-nnn}, where \(\mathcal{G}_L\) is a linear operator parameterized by \(\theta\). {\response This baseline is chosen to empirically justify the usage of deep neural operators for nudging.} Two metrics are under our consideration to assess the performance: the normalized root mean squared error (NRMSE) and its time-averaged version (aNRMSE), defined as follows:
\[
    \mathrm{NRMSE}(t_k) =  \frac{\|\hat{u}(t_k) - u(t_k)\|_2}{\|u(t_k)\|_2}, \qquad
    \mathrm{aNRMSE} = \frac{1}{K} \sum_{k=1}^K \mathrm{NRMSE}(t_k).
\]
Here, \(u(t_k)\) is the true state at time step \(t_k\), \(\hat{u}(t_k)\) is the estimated state, \(N\) is the number of spatial grid points, and \(K\) is the total number of time steps over which the error is averaged. We report NRMSE and aNRMSE when we are interested in errors in time and total error, respectively.

All data are generated in double precision, while model training and evaluation are performed in single precision. The implementation is based on \texttt{JAX}~\citep{jax2018github}, together with the following supporting libraries: \texttt{Equinox}~\citep{kidger2021equinox} for defining neural network modules, \texttt{JAXopt}~\citep{blondel2021jaxopt} for differentiable optimization routines, \texttt{Optax}~\citep{deepmind2020jax} for first-order optimization algorithms, and \texttt{Makie.jl}~\citep{DanischKrumbiegel2021} for visualization. Architectural details and hyperparameter settings are provided in \ref{app-detail}. The code is available at \url{https://github.com/jaeminoh/nnn}.

\subsection{The Lorenz 96 model}\label{sec-L96}
The Lorenz 96 model was introduced by \citet{lorenz1998optimal} as a system of ordinary differential equations designed to simulate an atmospheric quantity on a latitude circle. The goal of the model was to investigate optimal strategies for collecting supplementary observational data. Due to its chaotic nature, it has become a widely used benchmark in the data assimilation community.

The \(N\)-dimensional Lorenz 96 model is given by the following equations:
\begin{equation}\label{eq-L96}
    \frac{\dd u_n}{\dd t} = (u_{n+1} - u_{n-2}) u_{n-1} - u_n + F,
\end{equation}
where \(n = 1, \dots, N\), with periodic domain: \(u_{-1} = u_{N-1}\), \(u_{N+1} = u_1\), and \(u_0 = u_N\). The quadratic, linear, and constant terms in the equation represent the dynamics of advection, diffusion, and external forcing, respectively. A visualization of the chaotic behavior of the model is provided in \Cref{fig-chaos}, illustrating how a small perturbation to the first component of the initial condition grows rapidly over time.\footnote{The Lyapunov time, which characterizes the timescale at which small perturbations grow by a factor of \(e\), is approximately 0.6 for the Lorenz 96 model~\citep{bocquet2024accurate}.}

\begin{figure}[ht]
    \centering
    \includegraphics[width=\linewidth]{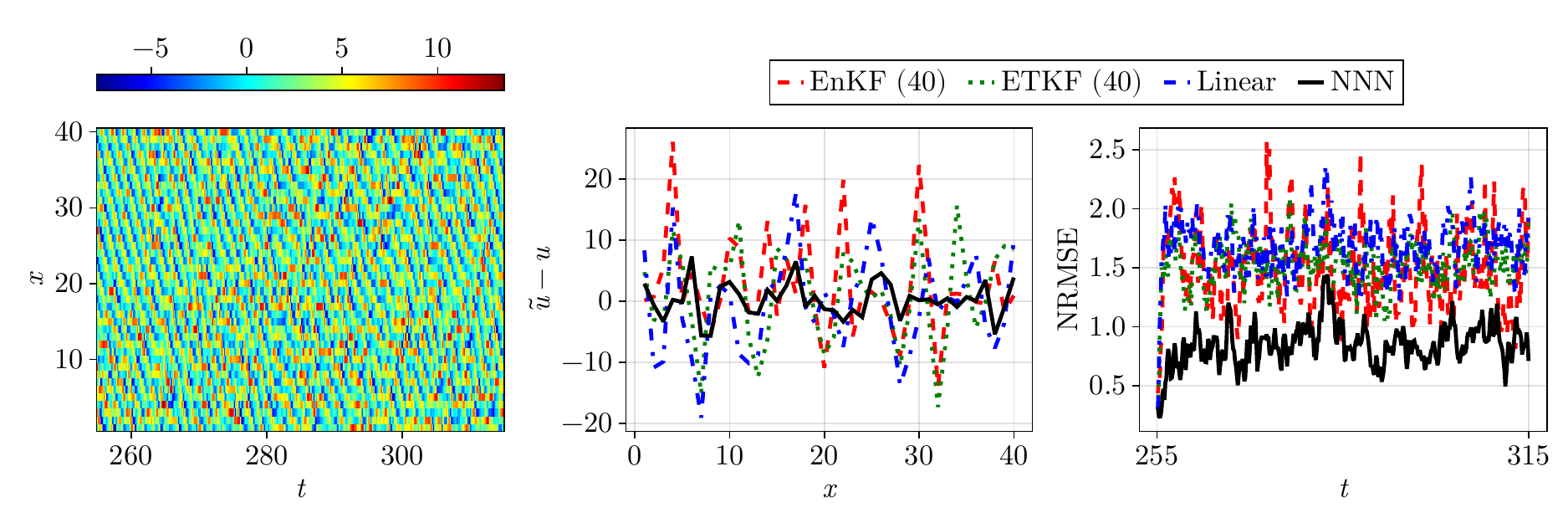}
    \caption{Visualization for the Lorenz 96 model with \(N=40\). The left panel presents the ground truth on the test time interval \([255, 315]\). The center and right panels visualize the performance comparison among EnKF, ETKF, Linear, and NNN with \(F=16\), \(\sigma=0.6298\), and 25\% observation. The center panel presents differences between analyses and ground truth at \(t = 315\), and the rightmost panel shows NRMSE growth in time.}
    \label{fig-L96}
\end{figure}

In our numerical experiments, we consider the case \(N = 40\) under three regimes:
\[
(F, \sigma) = (4, 0.1854), \quad (F, \sigma) = (8, 0.3640), \quad (F, \sigma) = (16, 0.6298).
\]
The system was integrated using the fourth-order Runge--Kutta method with a time step of \(0.01\), starting from the initial condition \((F + 0.01, F, \dots, F)^T\) and proceeding up to a final time \(t_f = 315\). Observations were generated according to
\[
    y(t) = \mathcal{H} u(t) + \sigma \varepsilon(t),
\]
where \(\mathcal{H}\) denotes a uniform subsampling operator. For example, for 50\% observations,
\[
\mathcal{H}_{50\%}(u_1, \dots, u_{2n})^T = (u_1, u_3, \dots, u_{2n-1})^T,
\]
and \(\varepsilon\) represents Gaussian noise. The time interval between consecutive observations was set to \(\Delta t = 0.15\).
{\response For training, we used the true trajectory for \(t \in [12, 255]\), with Gaussian noise added to the initial conditions. The cutoff of \(t\in [0, 12)\) is to exclude transient behavior from the training dataset.} During testing, we compared the true trajectory over \(t \in (255, 315]\) with the analysis trajectory initialized from the perturbed initial condition. The left panel of \cref{fig-L96} shows the solution of \cref{eq-L96} for \(F = 16\) on the test time period.

To evaluate the performance of NNN, we compare it against EnKF and ETKF using ensemble sizes of $20$ and $40$, as well as linear nudging. The forcing parameter $F$, observation noise level $\sigma$, and observation ratio are systematically varied. Quantitative results are reported in \cref{tab-L96-40}, where bold and underlined values indicate the best and second-best performance for each configuration, respectively.

Overall, NNN consistently achieves either the best or second-best accuracy across all tested settings. 
Its advantage is particularly evident in the challenging $F=16$ regime, where ensemble-based methods struggle under sparse observation conditions. When the observation ratio falls below 100\%, EnKF often fails to reduce the error effectively and frequently diverges, especially with the smaller ensemble size. 
While EnKF with $40$ members performs competitively under full observation, its stability deteriorates as sparsity increases. In contrast, ETKF and the data-driven nudging methods (Linear and NNN) remain stable across all cases. 
Notably, NNN consistently outperforms the other methods in all \(F=16\) configurations and maintains strong accuracy even at \(25\%\) observation. These findings are consistent with \citet{bocquet2024accurate}, who reported that machine learning–based data assimilation methods can achieve performance comparable to well-tuned ensemble Kalman filters without relying on ensemble simulations.

\begin{table}[t]
    \centering
    \caption{Quantitative comparison of data assimilation methods on the 40-dimensional Lorenz 96 system under varying forcing strength \(F\), observation noise level \(\sigma\), and observation sparsity. Results are reported in terms of aNRMSE. EnKF and ETKF are evaluated with ensemble sizes of \(20\) and \(40\). Boldface and underline indicate the best and second-best performance for each configuration, respectively.}
    \resizebox{\columnwidth}{!}{
    \begin{tabular}{c|ccc|ccc|ccc}
    \toprule
    \multirow{2}{*}{Method} & \multicolumn{3}{c|}{\(F=4,\ \sigma=0.1854\)} & \multicolumn{3}{c|}{\(F=8,\ \sigma=0.3640\)} & \multicolumn{3}{c}{\(F=16,\ \sigma=0.6298\)} \\
    \cline{2-10}
    & 100\% & 50\% & 25\% & 100\% & 50\% & 25\% & 100\% & 50\% & 25\% \\
    \hline
    EnKF (20) & 3.24E-1 & NaN & NaN & 1.68E-1 & NaN & NaN & 7.73E-1 & NaN & NaN \\
    ETKF (20) & 3.70E-1 & 1.01E0 & 1.14E0 & 4.01E-1 & 8.27E-1 & 1.02E0 & 1.22E0 & 1.44E0 & 1.56E0\\
    EnKF (40) & {\bf 6.91E-2} & NaN & 1.00E0 & 7.72E-2 & {\bf 2.45E-1} & {\bf 8.37E-1} & \underline{2.93E-1} & NaN & \underline{1.48E0}\\
    ETKF (40) & 1.56E-1 & \underline{7.15E-1} & \underline{9.22E-1} & 1.30E-1 & 6.32E-1 & 9.11E-1 & 1.13E0 & 1.39E0 & 1.53E0\\
    Linear & 1.01E0 & 1.01E0 & 1.03E0 & 1.09E0 & 1.13E0 & 1.17E0 & 1.49E0 & 1.59E0 & 1.69E0 \\
    \hline
    NNN (ours) & \underline{8.11E-2} & {\bf 2.23E-1} & {\bf 8.83E-1} & {\bf 7.20E-2} & \underline{2.47E-1} & \underline{8.60E-1} & {\bf 1.95E-1} & {\bf 3.87E-1} & {\bf 8.57E-1} \\
    \bottomrule
    \end{tabular}
    }
    \label{tab-L96-40}
\end{table}

\Cref{fig-L96} illustrates the numerical results. As observed, the system evolves rapidly and does not settle into a stationary state. The center and right panels compare the performance of EnKF, ETKF, Linear, and NNN for \((F, \sigma) = (16, 0.6298)\) with 25\% observations, representing the most challenging configuration considered. NNN more accurately tracks the ground truth and maintains a consistently lower NRMSE over time than the other methods.

\subsection{Kuramoto--Sivashinsky equation}\label{sec-KS}
The Kuramoto--Sivashinsky equation~\citep{kuramoto1976persistent,sivashinsky1980flame} was introduced in the mid-1970s in the study of reaction--diffusion systems and flame front stability~\citep{hyman1986kuramoto}. The equation reads
\begin{equation}\label{eq-kursiv}
    \partial_t u + u \partial_x u + \partial_x^2 u + \partial_x^4 u = 0,
\end{equation}
and is defined on the periodic domain \([0, 32\pi)\). The positive sign of the second-order diffusion term injects energy into the system, while the nonlinear advection term transfers energy from low to high wavenumbers. The fourth-order diffusion term subsequently stabilizes the system. 

A notable feature of the Kuramoto--Sivashinsky equation is its chaotic behavior. To illustrate this, we consider a smooth, slowly varying initial condition
\[
    u(x,0) = \cos(x/16) \bigl(1 + \sin(x/16)\bigr).
\]
As illustrated in \citet{kassam2005fourth}, the solution begins to exhibit chaotic dynamics around \(t = 50\). It is known that the Kuramoto--Sivashinsky equation possesses an inertial manifold that exponentially attracts all initial conditions~\citep{foias1988inertial,robinson1994inertial}, explaining why the trajectory from a simple initial condition rapidly transitions into chaotic behavior.

\begin{figure}[ht]
    \centering
    \includegraphics[width=\linewidth]{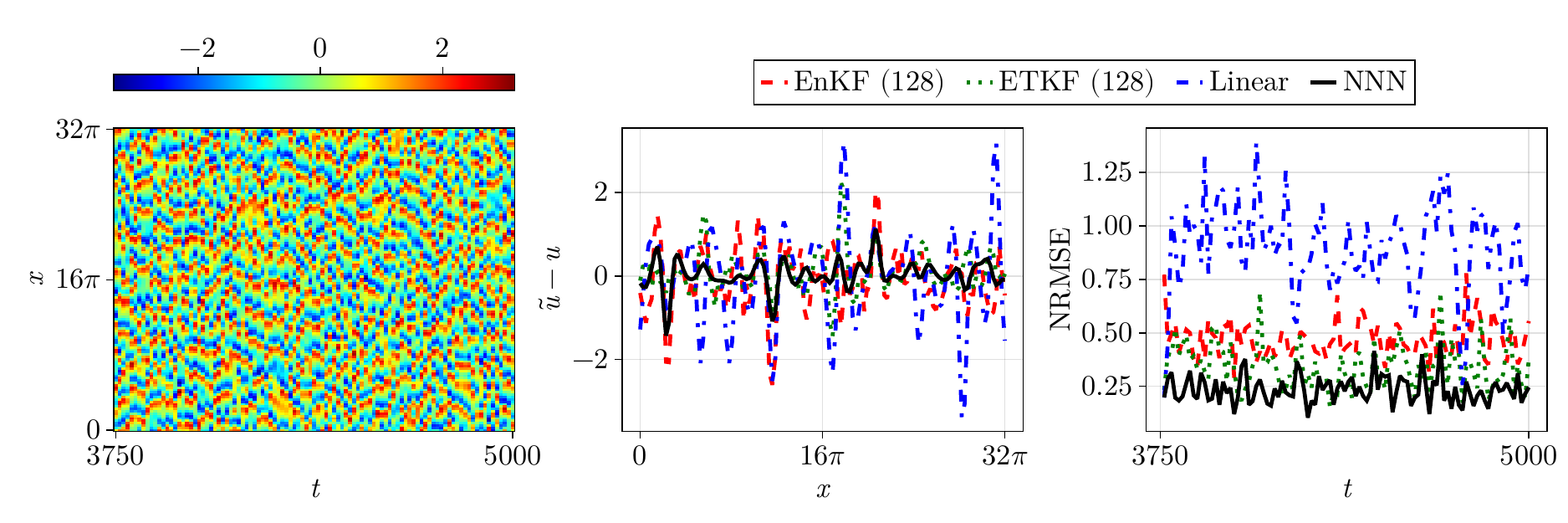}
    \caption{Visualization for the Kuramoto--Sivashinsky equation. The left panel shows the ground truth from \(t = 3750\) to \(t = 5000\), which is the test case.
    The center and right panels display qualitative performance comparison among EnKF, ETKF, Linear, and NNN with \(\sigma=0.5\) and \(25\%\) observations. The center panel shows differences between analyses and ground truth at \(t = 5000\), and the right panel shows NRMSE growth on the test time interval.}
    \label{fig-kursiv}
\end{figure}

To compute the ground truth, we employed the Fourier pseudo-spectral method with 128 modes and advanced in time using the fourth-order exponential time differencing Runge--Kutta scheme~\citep{kassam2005fourth} with a time step of \(0.025\). We set \(\Delta t = 0.25\).
{\response 
We integrated the system up to \(t = 5000\). The training dataset consists of snapshots for \(t\in[1250, 3750]\), and the test dataset consists of snapshots for \(t \in [3750, 5000]\). The left panel of \cref{fig-kursiv} illustrates ground truth solutions on the test time interval.
}

\Cref{tab-KS} reports quantitative comparisons between NNN, Linear, and ensemble-based Kalman filters for the Kuramoto--Sivashinsky equation under varying observation sparsity and noise levels. In contrast to the Lorenz 96 case (\cref{sec-L96}), ETKF generally outperforms EnKF across all configurations, particularly at higher noise levels and increased observation sparsity. 
This behavior can be attributed to ensemble-based filters' stochastic perturbations to enhance ensemble spread and improve error covariance estimation. Since the KS dynamics evolve on a smaller amplitude scale ($O(1)$) than those of the Lorenz 96 ($O(10)$), perturbations of the same magnitude may be ill-suited for EnKF, thereby favoring the superior accuracy of ETKF in this context.
Despite the improved performance of ETKF, NNN consistently achieves the lowest error across almost all configurations. The exceptions occur under full observation, where linear nudging marginally outperformed NNN -- the improvements are less than \(5\%\). Notably, as observations become sparse, NNN exhibits a clear advantage over both ensemble-based filters and linear nudging, maintaining robust accuracy even at \(25\%\) observations.

\begin{table}[t]
    \centering
    \caption{Quantitative comparison of data assimilation methods on the Kuramoto--Sivashinsky equation under varying observation noise levels \(\sigma\) and observation sparsity. Results are reported in terms of aNRMSE. EnKF and ETKF are evaluated with ensemble sizes of \(64\) and \(128\). Boldface and underline indicate the best and second-best performance for each configuration, respectively.}
    \resizebox{\columnwidth}{!}{
    \begin{tabular}{c|ccc|ccc|ccc}
    \toprule
    \multirow{2}{*}{Method} & \multicolumn{3}{c|}{\(\sigma=0.25\)} & \multicolumn{3}{c|}{\(\sigma=0.375\)} & \multicolumn{3}{c}{\(\sigma=0.5\)} \\
    \cline{2-10}
    & 100\% & 50\% & 25\% & 100\% & 50\% & 25\% & 100\% & 50\% & 25\% \\
    \hline
    EnKF (64) & 1.65E-1 & 3.58E-1 & 4.92E-1 & 2.21E-1 & 3.73E-1 & 5.52E-1 & 2.73E-1 & 4.01E-1 & 5.99E-1\\
    ETKF (64) & 1.13E-1 & 1.44E-1 & 2.62E-1 & 1.40E-1 & 1.70E-1 & 3.30E-1 & 1.60E-1 & 2.01E-1 & 4.39E-1 \\
    EnKF (128) & 1.64E-1 & 2.17E-1 & 3.22E-1 & 2.26E-1 & 2.77E-1 & 4.05E-1 & 2.77E-1 & 3.25E-1 & 4.68E-1 \\
    ETKF (128) & 1.11E-1 & 1.21E-1 & \underline{1.91E-1} & 1.30E-1 & 1.40E-1 & \underline{2.54E-1} & 1.41E-1 & \underline{1.62E-1} & \underline{3.44E-1} \\
    Linear & {\bf 5.41E-2} & \underline{1.02E-1} & 8.96E-1 & {\bf 8.09E-2} & \underline{1.48E-1} & 8.77E-1 & {\bf 1.08E-1} & 1.89E-1 & 8.85E-1 \\
    \hline
    NNN (ours) & \underline{5.51E-2} & {\bf 6.42E-2} & {\bf 1.28E-1} & \underline{8.26E-2} & {\bf 9.52E-2} & {\bf 1.78E-1} & \underline{1.09E-1} & {\bf 1.26E-1} & {\bf 2.23E-1} \\
    \bottomrule
    \end{tabular}
    }
    \label{tab-KS}
\end{table}

The center and right panels of \Cref{fig-kursiv} show qualitative comparisons of EnKF, ETKF, Linear, and NNN on \(\sigma = 0.5\) and 25\% observation, which is the most challenging case in the experiments. 
The center panel displays the pointwise differences between the estimated and ground-truth states at the final time step $t = 5000$. Evidently, the magnitude of the pointwise errors for NNN is the smallest, with ETKF following as a close second. 
The right panel depicts the evolution of NRMSE over time. Consistent with the previous results, NNN accurately tracks the ground-truth states, maintaining bounded NRMSE values throughout the test period. In contrast, the linear nudging method rapidly diverges from the true trajectory, underscoring the importance of employing neural networks to design effective data-driven nudging terms.

\subsection{Kolmogorov flow}\label{sec-KF}
In this final experiment, we consider a two-dimensional incompressible Navier-Stokes equation with Kolmogorov forcing~\citep{arnold1958seminar,chandler2013invariant}\footnote{\response For a rigorous mathematical treatment of nudging for the Navier--Stokes equation, we refer the reader to \citep{carlson2024determining}.}:
\[
\begin{aligned}
    \partial_t {\bf u} + ({\bf u} \cdot \nabla){\bf u} + \nabla p &= \nu \nabla^2 {\bf u} + {\bf f}, \\
    \nabla \cdot {\bf u} &= 0, \\
    {\bf f} &= \sin(4y)\begin{bmatrix}1 \\ 0 \end{bmatrix} -0.1 {\bf u},
\end{aligned}
\]
with viscosity \(\nu = 10^{-2}\), and step size \(7.01 \times 10^{-3}\) computed based on the CFL condition. Spatial domain is \([0, 2\pi)^2\) with periodic boundary. This system generates a statistically consistent turbulent flow, with the complexity of the flow controlled solely by \(\nu\). For the reference solver, we used a Fourier pseudo-spectral spatial discretization method with \(N^2 = 64^2\) with an implicit-explicit time stepper (Crank--Nicolson RK4) implemented in JAX-CFD~\citep{kochkov2021machine}. We employed \(K>1\) since the package is implemented in JAX.

{\response
To generate a dataset for training and testing the NNN, we follow the procedure described in \citep{frerix2021variational}: integrating a randomly sampled velocity field into a statistically stationary regime. 
We set observations to be made every \(\Delta t = 7.01 \times 10^{-2}\).
The time interval for the training dataset is approximately \([70.1, 210.4]\).
The time interval for the test dataset is approximately \([771.4, 841.5]\) to evaluate temporal generalization.}

In this benchmark problem, the state dimension is \(64^2 = 4096\), making it a significantly more challenging large-scale test case. Owing to the fully serial implementation of EnKF in FilterPy,\footnote{\url{https://github.com/rlabbe/filterpy}} we do not report EnKF results for this example. Instead, we compare ETKF with ensemble sizes of \(100\) and \(1000\). Quantitative results are summarized in \Cref{tab-kolmogorov}. In contrast to the Kuramoto--Sivashinsky equation, linear nudging performs poorly across all noise levels and observation rates, indicating limited scalability in terms of accuracy to high-dimensional, two-dimensional PDE systems. While increasing the ETKF ensemble size substantially improves performance under dense observations, its accuracy degrades rapidly as observations become sparse. In contrast, NNN consistently achieves the lowest errors across nearly all configurations. In particular, under highly sparse observations (\(6.25\%\)), NNN attains errors that are more than an order of magnitude smaller than those of ETKF and linear nudging, demonstrating strong robustness to observation sparsity and excellent scalability to large-scale two-dimensional dynamics.

\begin{table}[t]
    \centering
    \caption{Quantitative comparison of data assimilation methods on the two-dimensional Kolmogorov flow with state dimension \(64^2 = 4096\), under varying observation noise levels \(\sigma\) and observation rates. Results are reported in terms of aNRMSE. Ensemble sizes of ETKF are \(100\) or \(1000\). Boldface and underline indicate the best and second-best performance for each configuration, respectively.}
    \resizebox{\columnwidth}{!}{
    \begin{tabular}{c|ccc|ccc|ccc}
    \toprule
    \multirow{2}{*}{Method} & \multicolumn{3}{c|}{\(\sigma=0.5\)} & \multicolumn{3}{c|}{\(\sigma=0.75\)} & \multicolumn{3}{c}{\(\sigma=1\)} \\
    \cline{2-10}
    & 100\% & 25\% & 6.25\% & 100\% & 25\% & 6.25\% & 100\% & 25\% & 6.25\% \\
    \hline
    ETKF (100) & 5.41E-1 & 8.43E-1 & 1.06E0 & 6.35E-1 & 8.77E-1 & 1.14E0 & 7.10E-1 & 1.01E0 & 1.13E0 \\
    ETKF (1000) & \underline{5.40E-2} & \underline{6.89E-2} & \underline{8.49E-1} & \underline{5.99E-2} & \underline{3.11E-1} & \underline{1.01E0} & {\bf 6.33E-2} & \underline{6.90E-1} & \underline{1.09E0} \\
    Linear & 1.01E0 & 1.06E0 & 1.08E0 & 9.63E-1 & 1.09E0 & 1.12E0 & 1.01E0 & 1.12E0 & 1.12E0 \\
    \hline
    NNN (ours) & {\bf 3.37E-2} & {\bf 2.93E-2} & {\bf 3.96E-2} & {\bf 5.08E-2} & {\bf 4.35E-2} & {\bf 6.00E-2} & \underline{6.88E-2} & {\bf 5.79E-2} & {\bf 7.97E-2} \\
    \bottomrule
    \end{tabular}
    }
    \label{tab-kolmogorov}
\end{table}

\begin{figure}[t]
    \centering
    \includegraphics[width=\linewidth]{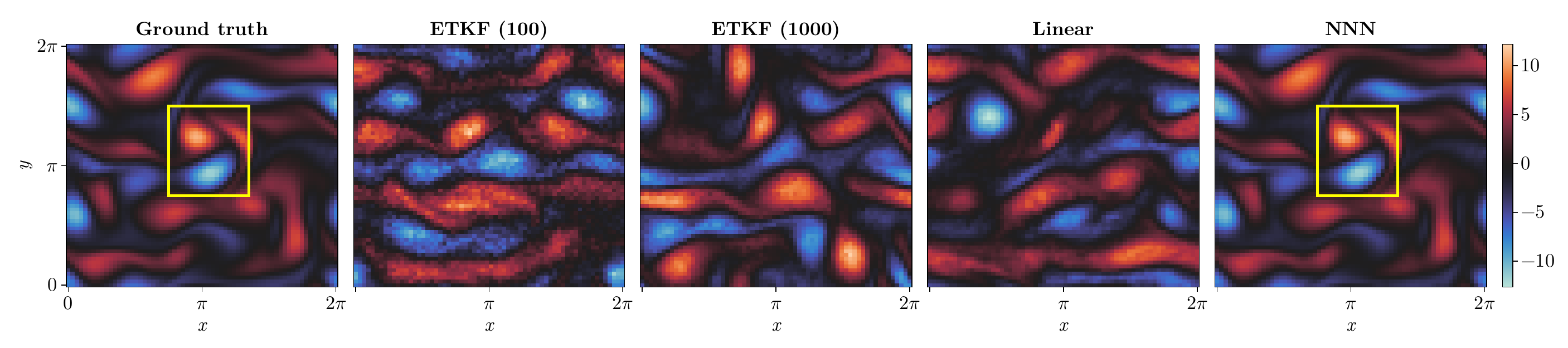}
    \caption{Ground truth and estimated states at the final test time \(t \approx 841\). The figure presents a qualitative performance comparison among ETKF with ensemble sizes 100 and 1000, Linear, and NNN for the test case of \(\sigma=1\) and \(6.25\%\) observations.}
    \label{fig-kolmogorov-snapshot}
\end{figure}

\begin{figure}[ht]
    \centering
    \includegraphics[width=0.75\linewidth]{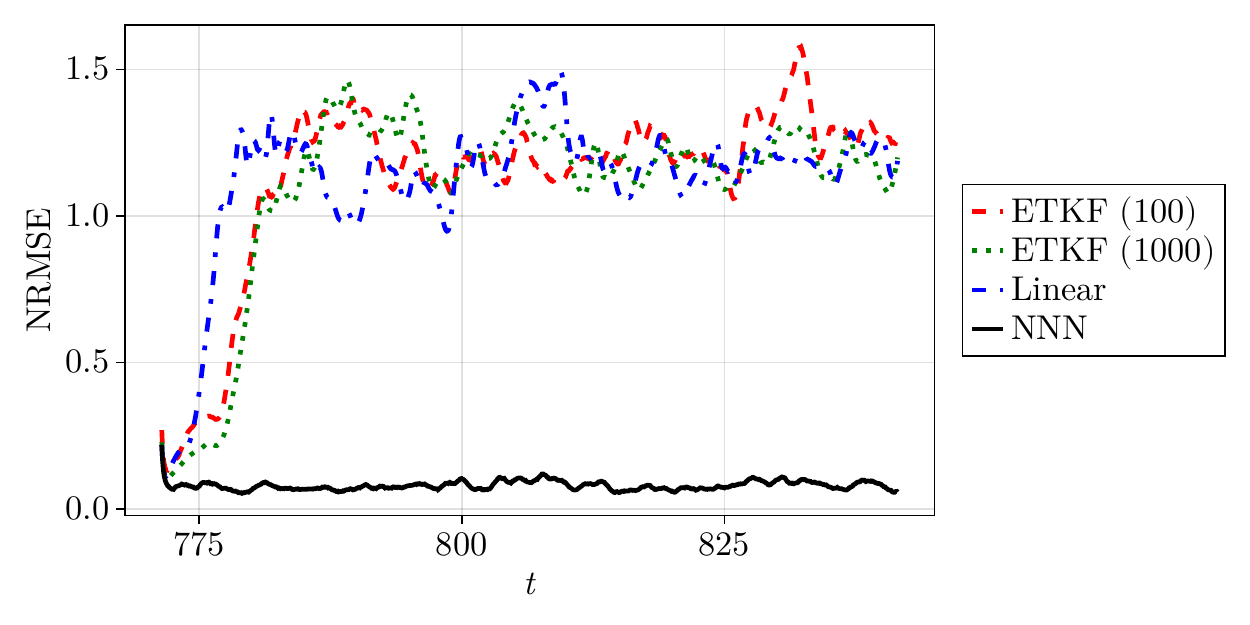}
    \caption{NRMSE growth on the test time interval \([771.4, 841.5]\) for the test case of \(\sigma = 1\) and \(6.25\%\) observations.}
    \label{fig-kolmogorov-nrmse}
\end{figure}

\Cref{fig-kolmogorov-snapshot} shows ground truth and estimated vorticities \(\omega = \tfrac{\partial {\bf u}_y}{\partial x} - \tfrac{\partial {\bf u}_x}{\partial y}\) at the final test time \(t\approx 841\) for the test case of \(\sigma = 1\) and 6.25\% observation, which is the most challenging case in the experiments. As highlighted with the yellow boxes, NNN successfully recovers a vorticity feature of the ground truth state, while the other methods completely fail. \Cref{fig-kolmogorov-nrmse} shows NRMSE values over the test time interval \(t \in [771, 841]\). The errors of NNN remain small, whereas the errors of the other methods grow rapidly.\footnote{\response \citet{buzzicotti2022inferring} also empirically showed supremacy of the machine learning approach over Bayesian inference approaches in the context of distinguishing different turbulent scenarios from data.}

{\response
\subsubsection{Accuracy on extended test time windows}
Kalman filter variants are independent of data and are therefore expected to maintain their accuracy regardless of the test time horizon. To determine whether NNN exhibits a similar level of temporal generalization, we evaluate its performance on test time windows that are longer than the training time window. We focus on the most challenging setting with observation noise level $\sigma = 1$ and an observation rate of $6.25\%$.

We first fix the training time interval to $[70.1, 210.4]$ and progressively extend the test interval from $[771.4, 841.5]$ to $[771.4, t_f]$, where \(t_f \in \{911.6,\,981.7,\,1051.9,\,1122.0\}\).
The corresponding aNRMSE values are reported in the left part of \cref{tab:kf-generalization}. Although the error increases slightly as the rollout horizon becomes longer, the aNRMSE remains below $0.09$ in all cases, indicating that NNN maintains stable prediction accuracy over substantially extended time windows.

Next, we fix the test interval to $[771.4, 1122.0]$ and investigate the effect of reducing the amount of training data. Specifically, the final time of the training interval is varied from $210.4$ to \(\{105.2,\,140.2,\,175.3\}\),
while keeping the initial time fixed at $70.1$. The right part of \cref{tab:kf-generalization} summarizes the resulting aNRMSE values. As expected, increasing the length of the training interval consistently improves prediction accuracy. Nevertheless, even with the shortest training interval, corresponding to only $500$ training snapshots, the aNRMSE remains below $0.09$. These results demonstrate that NNN generalizes well to long rollout horizons and retains strong predictive performance even when trained with relatively limited temporal data.

\begin{table}[t]
    \centering
    \caption{Accuracy of NNN under extended test horizons and varying training intervals.}
    \begin{tabular}{ccc|ccc}
    \toprule
    Train interval & Test interval & aNRMSE & Train interval & Test interval & aNRMSE \\
    \midrule
    \multirow{4}{*}{$[70.1,\,210.4]$}
        & $[771.4,\,911.6]$  & 8.10E-2 &
    $[70.1,\,105.2]$ & \multirow{4}{*}{$[771.4,\,1122.0]$} & 8.98E-2 \\
        & $[771.4,\,981.7]$  & 8.23E-2 &
    $[70.1,\,140.2]$ & & 8.58E-2 \\
        & $[771.4,\,1051.9]$ & 8.09E-2 &
    $[70.1,\,175.3]$ & & 8.23E-2 \\
        & $[771.4,\,1122.0]$ & 8.18E-2 &
    $[70.1,\,210.4]$ & & 8.18E-2 \\
    \bottomrule
    \end{tabular}
    \label{tab:kf-generalization}
\end{table}
}

\subsection{Computational cost}
In this section, we evaluate the computational efficiency of the ensemble-based Kalman filters and the proposed NNN. For EnKF, we employed the FilterPy library, which propagates ensemble members in a strictly serial manner and is therefore not suited for GPU acceleration. To establish a more competitive baseline, we implemented ETKF in JAX, leveraging \texttt{jax.vmap} to vectorize the analysis step across the ensemble dimension. All GPU-based experiments were conducted on a single NVIDIA A6000.

\Cref{tab-speed-comparison} reports the assimilation throughput (iterations per second) for each method across benchmark problems. The values \(N\) in parentheses denote the ensemble sizes used for EnKF and ETKF. For one-dimensional systems (Lorenz--96 and Kuramoto--Sivashinsky), the JAX-based ETKF implementation is more than \(5\times\) faster than the serial EnKF in FilterPy, reflecting the benefit of GPU vectorization. Across all benchmarks, both NNN and linear nudging consistently achieve higher throughput than ETKF. The training cost of Linear and NNN is negligible -- typically less than two minutes -- and is therefore excluded from the comparison. The largest speedup is observed for the two-dimensional Kolmogorov flow, where NNN attains approximately a \(25\times\) improvement over ETKF.

Combined with the superior accuracy of NNN over linear nudging (\cref{sec-KF}), these results point to the favorable scalability of NNN for high-dimensional systems.

\begin{table}[!htp]\centering
\caption{Assimilation throughput (iterations per second) for different methods across benchmark problems. State dimensions \(N\) are indicated in parentheses. Ensemble sizes were set to match the state dimensions. Reported timings exclude training costs for Linear and NNN. The fully serial EnKF implementation exhibits poor scalability with increasing state dimension, whereas the JAX-based ETKF benefits from GPU vectorization. The higher throughput of Linear and NNN demonstrates that, once trained, data-driven nudging can substantially reduce the computational cost of data assimilation.}
\begin{tabular}{l|ccc}\toprule
& L96 (\(N=40\)) & KS (\(N=128\)) & KF (\(N=1000\)) \\
\midrule
EnKF (single CPU) & 49.07it/s & 4.70it/s & 0.0069it/s\\
ETKF (GPU) & 1026.49it/s & 409.91it/s & 8.24it/s\\
Linear (GPU) & 1855.03it/s & 893.26it/s & 222.14it/s \\
NNN (GPU) & 1279.08it/s & 814.98it/s & 225.90it/s \\
\bottomrule
\end{tabular}
\label{tab-speed-comparison}
\end{table}

\section{Conclusion}\label{sec-conclusion}
In this work, we have explored NNN, a neural operator-based, data-driven approach for designing nudging terms in nonlinear state-space models. The key idea is to represent the nudging term using deep neural operators. Once trained, data assimilation reduces to a simple forward pass through the network, which is computationally efficient, particularly on GPUs, as it avoids matrix inversion and ensemble simulations—two primary computational bottlenecks in traditional methods. The training cost is incurred only once, making the method highly efficient during the operational phase. We have also provided a theoretical existence result grounded in the KKL observer theory. Finally, we assessed the performance of NNN across three benchmark problems: the Lorenz 96 model, the Kuramoto--Sivashinsky equation, and the Kolmogorov flow. NNN achieved the first-to-second best performance across all test cases and demonstrated robustness in handling sparse and noisy observations.

Despite its promise, our approach has several limitations that warrant further investigation. This work serves primarily as a proof of concept, and we have not yet explored realistic applications. While we chose chaotic dynamical systems to highlight challenges relevant to numerical weather prediction, we have not tested the scalability of our method on more complex two- or three-dimensional physical domains or its robustness in the presence of model error. Addressing these aspects remains an important direction for future work. Moreover, we have assumed access to ground truth states, an assumption that often does not hold in realistic applications. In practice, reanalysis datasets~\citep{hersbach2020era5} may be utilized, but developing approaches that remove reliance on ground truth data would be a valuable research direction. {\response On the theoretical side, extending \Cref{thm-uat-kkl} to account for general noisy observations~\citep{sanz2015long,oljaca2018almost}, thereby better reflecting realistic scenarios, or developing a mathematical framework that relaxes existing assumptions while providing error rates~\citep{olson2003determining, azouani2014continuous, carlson2026nudging}, would also be promising directions for future research.}

We believe that this work stacks a contribution towards scalable, efficient, and accurate data assimilation methods in high-dimensional chaotic systems, moving toward practical applications in numerical weather prediction and beyond.

\section*{Acknowledgement}
The first author thanks Sungho Han and Jinsol Seo for their comments with \Cref{thm-uat-kkl}. The authors thank two anonymous reviewers for their insightful comments.
The work of Y. Hong was supported by the Korea government (MSIT) through the NRF (RS-2023-00219980 \& RS-2026-25491270); and by the Institute of Information \& Communications Technology Planning \& Evaluation (IITP) grant funded by the Korea government (MSIT) (RS-2021-II211343, Artificial Intelligence Graduate School Program, Seoul National University).
The authors would like to express their
sincere gratitude to the reviewer for his/her insightful comments and suggestions.

\appendix

\section{Neural networks and hyperparameters}\label{app-detail}
\paragraph{Multi-layer perceptrons}
An \(L\)-layer perceptron (\(\mathrm{MLP}:\mathbb{R}^{d_0} \ni x \mapsto y \in\mathbb{R}^{d_L}\)) of an architecture \((d_0, d_1, \dots, d_L)\) and an activation function \(\phi\) is defined as follows:
\begin{align*}
    h_0 &= x, \\
    h_i &= \phi(W^ih_{i-1} + b^i), \qquad i=1, \dots, L-1 \\
    y &= W^L h_{L-1} + b^L,
\end{align*}
where \(\phi\) is applied element-wise, \(W^i \in \mathbb{R}^{d_i \times d_{i-1}}\) and \(b^i \in \mathbb{R}^{d_i}\). We call \(\theta = \{(W^i, b^i)|i=1, \dots, L\}\) as network parameters.

\paragraph{Convolutional layers}
Throughout the experiments, we have used \(\texttt{stride}=1\), \(\texttt{kernel size}=5\), \(\texttt{padding}=\texttt{circular}\), and \(\texttt{padding mode}=\texttt{same}\). Then we have a convolution kernel \(W \in \mathbb{R}^{C_{out} \times 5}\) and a bias \(b \in \mathbb{R}^{C_{out} \times N}\) through which we optimize appropriate loss functions (\cref{eq-loss}). The one-dimensional convolutional layer
\[
    \mathrm{Conv1d}: \mathbb{R}^{C_\mathrm{in} \times N} \rightarrow \mathbb{R}^{C_\mathrm{out} \times N}
\]
is defined by
\[
    \mathrm{Conv1d}(X)[i, j] = b[i, j] + \sum_{c=1}^{C_\mathrm{in}}\sum_{k=-2}^{2} W[i, k] X[c, j+k],
\]
where the \(-2\)-based indexing of the kernel was from \(\texttt{same}\) padding mode, and \(\texttt{circular}\) padding means \(X[:, j] = X[:, j \mod N]\). The two-dimensional extension is straightforward. For transposed convolutional layers, please see~\citep{dumoulin2016guide}.

\paragraph{Deep neural operator}
\[
    \mathcal{G}_\theta: (\tilde u, y) \mapsto \sum_{c=1}^C \mathcal{B}_c(\tilde u)  \mathcal{T}_c(y - H(\tilde u)),
\]
where the trunk network is defined by
\[
    \mathrm{Conv} \circ \tanh \circ\ \mathrm{ConvTransposed}: \mathbb{R}^{1 \times p} \ni y - H(\tilde u) \mapsto \mathcal{T} \in \mathbb{R}^{C \times d},
\]
and the branch network is defined as a two-layer MLP with \(\tanh\) activation function,
\[
    \mathrm{MLP}: \mathbb{R}^{d} \ni \tilde u \mapsto \mathcal{B} \in \mathbb{R}^C.
\]
We have taken this form since the branch and trunk networks serve the roles of coefficients and basis functions, respectively. Also, the number of channels, \(C\), can be thought of as the number of basis functions.

\begin{table}[ht]
    \centering
    \begin{tabular}{|c|c|c|c|}
        \hline
        & L96 (Section~\ref{sec-L96}) & KS (Section~\ref{sec-KS}) & KF (Section~\ref{sec-KF}) \\ \hline
        Epoch & 300 & 300 & 100 \\ \hline
        Learning rate & \multicolumn{3}{c|}{\(10^{-3}\)} \\ \hline
        Hidden channels & \multicolumn{3}{c|}{20} \\ \hline
        kernel size & \multicolumn{3}{c|}{5} \\ \hline
        stride & \multicolumn{3}{c|}{1} \\ \hline
        Burn-in steps & 80 & 5000 & 1000 \\ \hline
        Training steps & 1620 & 10000 & 2000 \\ \hline
        Test steps & 400 & 5000 & 1000 \\ \hline
        \(\Delta t\) & 0.15 & 0.25 & 7.01E-2 \\ \hline
        \texttt{inner\_steps} & 15 & 10 & 10 \\ \hline
        \(K\) & 5 & 10 &  10 \\ \hline
    \end{tabular}
    \caption{Configurations for three benchmark problems.}
    \label{tab-training-detail}
\end{table}

\section{Additional tables and figures}
\Cref{tab-K-study} presents the effect of \(K>1\) across {\response the} three benchmark problems considered in \Cref{sec-results}. Clearly, \(K>1\) achieved lower RMSE and nRMSE values than \(K=1\).
\begin{table}[ht]
    \centering
    \begin{tabular}{|c|c|c|c|}
    \hline
         & L96& KS& KF\\
     \hline
         \(K=1\)& 6.18E-1 & 1.67E-1 & 1.50E-1 \\
     \hline
         \(K>1\)& 3.46E-1 & 5.85E-2 & 1.17E-1 \\
     \hline
    \end{tabular}
    \caption{The effect of  different \(K\). The other experimental configurations correspond to the upper left slots of \Cref{tab-L96-40,tab-KS,tab-kolmogorov}.}
    \label{tab-K-study}
\end{table}

\bibliographystyle{plainnat}
\bibliography{references}

@article{buzzicotti2022inferring,
  title={Inferring turbulent environments via machine learning},
  author={Buzzicotti, Michele and Bonaccorso, Fabio},
  journal={The European Physical Journal E},
  volume={45},
  number={12},
  pages={102},
  year={2022},
  publisher={Springer}
}

@article{oljaca2018almost,
  title={Almost sure error bounds for data assimilation in dissipative systems with unbounded observation noise},
  author={Oljaca, Lea and Brocker, Jochen and Kuna, Tobias},
  journal={SIAM Journal on Applied Dynamical Systems},
  volume={17},
  number={4},
  pages={2882--2914},
  year={2018},
  publisher={SIAM}
}

@article{sanz2015long,
  title={Long-time asymptotics of the filtering distribution for partially observed chaotic dynamical systems},
  author={Sanz-Alonso, Daniel and Stuart, Andrew M},
  journal={SIAM/ASA Journal on Uncertainty Quantification},
  volume={3},
  number={1},
  pages={1200--1220},
  year={2015},
  publisher={SIAM}
}

@article{carlson2026nudging,
  title={On the nudging approach to continuous data assimilation in the limit of infinite error feedback gain},
  author={Carlson, Elizabeth and Farhat, Aseel and Martinez, Vincent R and Victor, Collin},
  journal={SIAM Journal on Control and Optimization},
  volume={64},
  number={3},
  pages={2026--2057},
  year={2026},
  publisher={SIAM}
}

@article{carlson2024determining,
  title={Determining modes, synchronization, and intertwinement},
  author={Carlson, Elizabeth and Farhat, Aseel and Martinez, Vincent R and Victor, Collin},
  journal={arXiv e-prints},
  pages={arXiv--2408},
  year={2024}
}

@article{azouani2014continuous,
  title={Continuous data assimilation using general interpolant observables},
  author={Azouani, Abderrahim and Olson, Eric and Titi, Edriss S},
  journal={Journal of Nonlinear Science},
  volume={24},
  number={2},
  pages={277--304},
  year={2014},
  publisher={Springer}
}

@article{olson2003determining,
  title={Determining modes for continuous data assimilation in 2D turbulence},
  author={Olson, Eric and Titi, Edriss S},
  journal={Journal of statistical physics},
  volume={113},
  number={5},
  pages={799--840},
  year={2003},
  publisher={Springer}
}

@article{oh2026effects,
  title={On the Effect of Neural Field Reparameterization for 4DVAR},
  author={Oh, Jaemin},
  journal={SIAM Journal on Scientific Computing},
  year={2026},
  publisher={SIAM},
  note={To appear}
}

@inproceedings{barthel2023machine,
  title={A machine learning framework for correcting under-resolved simulations of turbulent systems using nudged datasets},
  author={Barthel, Benedikt and Sapsis, Themis},
  booktitle={NeurIPS 2023 Workshop on Tackling Climate Change with Machine Learning},
  url={https://www.climatechange.ai/papers/neurips2023/21},
  year={2023}
}

@article{kassam2005fourth,
  title={Fourth-order time-stepping for stiff PDEs},
  author={Kassam, Aly-Khan and Trefethen, Lloyd N},
  journal={SIAM Journal on Scientific Computing},
  volume={26},
  number={4},
  pages={1214--1233},
  year={2005},
  publisher={SIAM}
}

@article{kochkov2021machine,
  title={Machine learning--accelerated computational fluid dynamics},
  author={Kochkov, Dmitrii and Smith, Jamie A and Alieva, Ayya and Wang, Qing and Brenner, Michael P and Hoyer, Stephan},
  journal={Proceedings of the National Academy of Sciences},
  volume={118},
  number={21},
  pages={e2101784118},
  year={2021},
  publisher={National Acad Sciences}
}

@article{lorenz1963deterministic,
  title={Deterministic nonperiodic flow},
  author={Lorenz, Edward N},
  journal={Journal of atmospheric sciences},
  volume={20},
  number={2},
  pages={130--141},
  year={1963}
}

@software{jax2018github,
  author = {James Bradbury and Roy Frostig and Peter Hawkins and Matthew James Johnson and Chris Leary and Dougal Maclaurin and George Necula and Adam Paszke and Jake Vander{P}las and Skye Wanderman-{M}ilne and Qiao Zhang},
  title = {{JAX}: composable transformations of {P}ython+{N}um{P}y programs},
  url = {http://github.com/jax-ml/jax},
  version = {0.3.13},
  year = {2018},
}

@inproceedings{filoche2023learning,
  title={Learning 4DVAR inversion directly from observations},
  author={Filoche, Arthur and Brajard, Julien and Charantonis, Anastase and B{\'e}r{\'e}ziat, Dominique},
  booktitle={International Conference on Computational Science},
  pages={414--421},
  year={2023},
  organization={Springer}
}

@article{bocquet2024accurate,
  title={Accurate deep learning-based filtering for chaotic dynamics by identifying instabilities without an ensemble},
  author={Bocquet, Marc and Farchi, Alban and Finn, Tobias S and Durand, Charlotte and Cheng, Sibo and Chen, Yumeng and Pasmans, Ivo and Carrassi, Alberto},
  journal={Chaos: An Interdisciplinary Journal of Nonlinear Science},
  volume={34},
  number={9},
  year={2024},
  publisher={AIP Publishing}
}

@book{law2015data,
  title={Data assimilation},
  author={Law, Kody and Stuart, Andrew},
  publisher={Springer},
  year={2015}
}

@article{chen1995universal,
  title={Universal approximation to nonlinear operators by neural networks with arbitrary activation functions and its application to dynamical systems},
  author={Chen, Tianping and Chen, Hong},
  journal={IEEE transactions on neural networks},
  volume={6},
  number={4},
  pages={911--917},
  year={1995},
  publisher={IEEE}
}

@inproceedings{frerix2021variational,
  title={Variational data assimilation with a learned inverse observation operator},
  author={Frerix, Thomas and Kochkov, Dmitrii and Smith, Jamie and Cremers, Daniel and Brenner, Michael and Hoyer, Stephan},
  booktitle={International Conference on Machine Learning},
  pages={3449--3458},
  year={2021},
  organization={PMLR}
}

@article{rozet2023score,
  title={Score-based data assimilation},
  author={Rozet, Fran{\c{c}}ois and Louppe, Gilles},
  journal={Advances in Neural Information Processing Systems},
  volume={36},
  pages={40521--40541},
  year={2023}
}

@article{kazantzis1998nonlinear,
  title={Nonlinear observer design using Lyapunov’s auxiliary theorem},
  author={Kazantzis, Nikolaos and Kravaris, Costas},
  journal={Systems \& Control Letters},
  volume={34},
  number={5},
  pages={241--247},
  year={1998},
  publisher={Elsevier}
}

@article{luenberger1964observing,
  title={Observing the state of a linear system},
  author={Luenberger, David G},
  journal={IEEE transactions on military electronics},
  volume={8},
  number={2},
  pages={74--80},
  year={1964},
  publisher={IEEE}
}

@article{bao2024score,
  title={A score-based filter for nonlinear data assimilation},
  author={Bao, Feng and Zhang, Zezhong and Zhang, Guannan},
  journal={Journal of Computational Physics},
  volume={514},
  pages={113207},
  year={2024},
  publisher={Elsevier}
}

@article{chandler2013invariant,
  title={Invariant recurrent solutions embedded in a turbulent two-dimensional Kolmogorov flow},
  author={Chandler, Gary J and Kerswell, Rich R},
  journal={Journal of Fluid Mechanics},
  volume={722},
  pages={554--595},
  year={2013},
  publisher={Cambridge University Press}
}

@book{leveque2002finite,
  title={Finite volume methods for hyperbolic problems},
  author={LeVeque, Randall J},
  volume={31},
  year={2002},
  publisher={Cambridge university press}
}

@book{hesthaven2007spectral,
  title={Spectral methods for time-dependent problems},
  author={Hesthaven, Jan S and Gottlieb, Sigal and Gottlieb, David},
  volume={21},
  year={2007},
  publisher={Cambridge University Press}
}

@book{trefethen1996finite,
  title={Finite difference and spectral methods for ordinary and partial differential equations},
  author={Trefethen, Lloyd Nicholas},
  year={1996},
  publisher={Cornell University-Department of Computer Science and Center for Applied~…}
}

@article{trefethen1993hydrodynamic,
  title={Hydrodynamic stability without eigenvalues},
  author={Trefethen, Lloyd N and Trefethen, Anne E and Reddy, Satish C and Driscoll, Tobin A},
  journal={Science},
  volume={261},
  number={5121},
  pages={578--584},
  year={1993},
  publisher={American Association for the Advancement of Science}
}

@article{clark2020synchronization,
  title={Synchronization to big data: Nudging the Navier-Stokes equations for data assimilation of turbulent flows},
  author={Clark Di Leoni, Patricio and Mazzino, Andrea and Biferale, Luca},
  journal={Physical Review X},
  volume={10},
  number={1},
  pages={011023},
  year={2020},
  publisher={APS}
}

@article{kidger2021equinox,
  title={Equinox: neural networks in JAX via callable PyTrees and filtered transformations},
  author={Kidger, Patrick and Garcia, Cristian},
  journal={arXiv preprint arXiv:2111.00254},
  year={2021}
}

@software{deepmind2020jax,
  title = {The {D}eep{M}ind {JAX} {E}cosystem},
  author = {DeepMind and Babuschkin, Igor and Baumli, Kate and Bell, Alison and Bhupatiraju, Surya and Bruce, Jake and Buchlovsky, Peter and Budden, David and Cai, Trevor and Clark, Aidan and Danihelka, Ivo and Dedieu, Antoine and Fantacci, Claudio and Godwin, Jonathan and Jones, Chris and Hemsley, Ross and Hennigan, Tom and Hessel, Matteo and Hou, Shaobo and Kapturowski, Steven and Keck, Thomas and Kemaev, Iurii and King, Michael and Kunesch, Markus and Martens, Lena and Merzic, Hamza and Mikulik, Vladimir and Norman, Tamara and Papamakarios, George and Quan, John and Ring, Roman and Ruiz, Francisco and Sanchez, Alvaro and Sartran, Laurent and Schneider, Rosalia and Sezener, Eren and Spencer, Stephen and Srinivasan, Srivatsan and Stanojevi\'{c}, Milo\v{s} and Stokowiec, Wojciech and Wang, Luyu and Zhou, Guangyao and Viola, Fabio},
  url = {http://github.com/google-deepmind},
  year = {2020},
}

@article{blondel2021jaxopt,
  title={Efficient and Modular Implicit Differentiation},
  author={Blondel, Mathieu and Berthet, Quentin and Cuturi, Marco and Frostig, Roy 
    and Hoyer, Stephan and Llinares-L{\'o}pez, Felipe and Pedregosa, Fabian 
    and Vert, Jean-Philippe},
  journal={arXiv preprint arXiv:2105.15183},
  year={2021}
}

@article{pareschi2000implicit,
  title={Implicit-explicit Runge-Kutta schemes for stiff systems of differential equations},
  author={Pareschi, Lorenzo and Russo, Giovanni and others},
  journal={Recent trends in numerical analysis},
  volume={3},
  pages={269--289},
  year={2000}
}

@article{DanischKrumbiegel2021,
  doi = {10.21105/joss.03349},
  url = {https://doi.org/10.21105/joss.03349},
  year = {2021},
  publisher = {The Open Journal},
  volume = {6},
  number = {65},
  pages = {3349},
  author = {Simon Danisch and Julius Krumbiegel},
  title = {{Makie.jl}: Flexible high-performance data visualization for {Julia}},
  journal = {Journal of Open Source Software}
}

@article{kim2024wearable,
author = {Kim, Dae Wook and Lee, Minki P. and Forger, Daniel B.},
title = {Wearable Data Assimilation to Estimate the Circadian Phase},
journal = {SIAM Journal on Applied Mathematics},
volume = {84},
number = {3},
pages = {S452-S475},
year = {2024},
doi = {10.1137/22M1509680},
URL = {https://doi.org/10.1137/22M1509680},
eprint = {https://doi.org/10.1137/22M1509680}
}

@article{lei2015nudging,
  title={Nudging, ensemble, and nudging ensembles for data assimilation in the presence of model error},
  author={Lei, Lili and Hacker, Joshua P},
  journal={Monthly Weather Review},
  volume={143},
  number={7},
  pages={2600--2610},
  year={2015}
}

@article{lorenz1998optimal,
  title={Optimal sites for supplementary weather observations: Simulation with a small model},
  author={Lorenz, Edward N and Emanuel, Kerry A},
  journal={Journal of the Atmospheric Sciences},
  volume={55},
  number={3},
  pages={399--414},
  year={1998}
}

@article{gilbert1984approach,
  title={An approach to nonlinear feedback control with applications to robotics},
  author={Gilbert, Elmer G and Ha, In Joong},
  journal={IEEE transactions on systems, man, and cybernetics},
  number={6},
  pages={879--884},
  year={1984},
  publisher={IEEE}
}

@article{robinson1994inertial,
  title={Inertial manifolds for the Kuramoto-Sivashinsky equation},
  author={Robinson, James C},
  journal={Physics Letters A},
  volume={184},
  number={2},
  pages={190--193},
  year={1994},
  publisher={Elsevier}
}

@article{foias1988inertial,
  title={Inertial manifolds for nonlinear evolutionary equations},
  author={Foias, Ciprian and Sell, George R and Temam, Roger},
  journal={Journal of differential equations},
  volume={73},
  number={2},
  pages={309--353},
  year={1988},
  publisher={Elsevier}
}

@book{lorenz1993essence,
  title={THE ESSENCE OF CHAOS},
  author={Lorenz, Edward N},
  year={1993}
}

@article{arnold1958seminar,
  title={The seminar of AN Kolmogorov on selected topics in analysis},
  author={Arnold, VI and Meshalkin, LD},
  journal={Usp. Mat. Nauk},
  volume={15},
  pages={247--250},
  year={1958}
}

@article{brivadis2023further,
  title={Further remarks on KKL observers},
  author={Brivadis, Lucas and Andrieu, Vincent and Bernard, Pauline and Serres, Ulysse},
  journal={Systems \& Control Letters},
  volume={172},
  pages={105429},
  year={2023},
  publisher={Elsevier}
}

@article{chen2018neural,
  title={Neural ordinary differential equations},
  author={Chen, Ricky TQ and Rubanova, Yulia and Bettencourt, Jesse and Duvenaud, David K},
  journal={Advances in neural information processing systems},
  volume={31},
  year={2018}
}

@article{antil2024ninns,
  title={NINNs: Nudging induced neural networks},
  author={Antil, Harbir and L{\"o}hner, Rainald and Price, Randy},
  journal={Physica D: Nonlinear Phenomena},
  volume={470},
  pages={134364},
  year={2024},
  publisher={Elsevier}
}

@incollection{antil2024data,
  title={Data assimilation with deep neural nets informed by nudging},
  author={Antil, Harbir and L{\"o}hner, Rainald and Price, Randy},
  booktitle={Reduction, Approximation, Machine Learning, Surrogates, Emulators and Simulators: RAMSES},
  pages={17--41},
  year={2024},
  publisher={Springer}
}

@phdthesis{kidger2021on,
    title={{O}n {N}eural {D}ifferential {E}quations},
    author={Patrick Kidger},
    year={2021},
    school={University of Oxford},
}

@article{dumoulin2016guide,
  title={A guide to convolution arithmetic for deep learning},
  author={Dumoulin, Vincent and Visin, Francesco},
  journal={arXiv preprint arXiv:1603.07285},
  year={2016}
}

@inproceedings{gordon1993novel,
  title={Novel approach to nonlinear/non-Gaussian Bayesian state estimation},
  author={Gordon, Neil J and Salmond, David J and Smith, Adrian FM},
  booktitle={IEE proceedings F (radar and signal processing)},
  volume={140},
  number={2},
  pages={107--113},
  year={1993},
  organization={IET}
}

@book{abidi1992data,
  title={Data fusion in robotics and machine intelligence},
  author={Abidi, Mongi A and Gonzalez, Rafael C},
  year={1992},
  publisher={academic press New York}
}

@book{nocedal1999numerical,
  title={Numerical optimization},
  author={Nocedal, Jorge and Wright, Stephen J},
  year={1999},
  publisher={Springer}
}

@book{bellman1957dynamic,
  title={Dynamic Programming},
  author={Bellman, R. and Rand Corporation and Karreman Mathematics Research Collection},
  isbn={9780691079516},
  lccn={57005444},
  series={Rand Corporation research study},
  url={https://books.google.com/books?id=wdtoPwAACAAJ},
  year={1957},
  publisher={Princeton University Press}
}

@article{luenberger1971introduction,
  title={An introduction to observers},
  author={Luenberger, David},
  journal={IEEE Transactions on automatic control},
  volume={16},
  number={6},
  pages={596--602},
  year={1971},
  publisher={IEEE}
}

@article{luenberger1966observers,
  title={Observers for multivariable systems},
  author={Luenberger, David},
  journal={IEEE transactions on automatic control},
  volume={11},
  number={2},
  pages={190--197},
  year={1966},
  publisher={IEEE}
}

@article{kuramoto1976persistent,
  title={Persistent propagation of concentration waves in dissipative media far from thermal equilibrium},
  author={Kuramoto, Yoshiki and Tsuzuki, Toshio},
  journal={Progress of theoretical physics},
  volume={55},
  number={2},
  pages={356--369},
  year={1976},
  publisher={Oxford University Press}
}

@article{sivashinsky1980flame,
  title={On flame propagation under conditions of stoichiometry},
  author={Sivashinsky, Gregory I},
  journal={SIAM Journal on Applied Mathematics},
  volume={39},
  number={1},
  pages={67--82},
  year={1980},
  publisher={SIAM}
}

@article{hyman1986kuramoto,
  title={The Kuramoto-Sivashinsky equation: a bridge between PDE's and dynamical systems},
  author={Hyman, James M and Nicolaenko, Basil},
  journal={Physica D: Nonlinear Phenomena},
  volume={18},
  number={1-3},
  pages={113--126},
  year={1986},
  publisher={Elsevier}
}

@article{hersbach2020era5,
  title={The ERA5 global reanalysis},
  author={Hersbach, Hans and Bell, Bill and Berrisford, Paul and Hirahara, Shoji and Hor{\'a}nyi, Andr{\'a}s and Mu{\~n}oz-Sabater, Joaqu{\'\i}n and Nicolas, Julien and Peubey, Carole and Radu, Raluca and Schepers, Dinand and others},
  journal={Quarterly journal of the royal meteorological society},
  volume={146},
  number={730},
  pages={1999--2049},
  year={2020},
  publisher={Wiley Online Library}
}

@article{carlson2024super,
  title={Super-Exponential Convergence Rate of a Nonlinear Continuous Data Assimilation Algorithm: The 2D Navier--Stokes Equation Paradigm},
  author={Carlson, Elizabeth and Larios, Adam and Titi, Edriss S},
  journal={Journal of Nonlinear Science},
  volume={34},
  number={2},
  pages={37},
  year={2024},
  publisher={Springer}
}

@article{lu2019deeponet,
  title={Deeponet: Learning nonlinear operators for identifying differential equations based on the universal approximation theorem of operators},
  author={Lu, Lu and Jin, Pengzhan and Karniadakis, George Em},
  journal={arXiv preprint arXiv:1910.03193},
  year={2019}
}

\end{document}